%% file: neurips_2025.tex
\documentclass{article}

\usepackage[final]{neurips_2025}

\usepackage{xcolor}         % colors

\usepackage[utf8]{inputenc} % allow utf-8 input
\usepackage[T1]{fontenc}    % use 8-bit T1 fonts
\usepackage{hyperref}       % hyperlinks
\usepackage{url}            % simple URL typesetting
\usepackage{booktabs}       % professional-quality tables
\usepackage{amsfonts}       % blackboard math symbols
\usepackage{nicefrac}       % compact symbols for 1/2, etc.
\usepackage{microtype}      % microtypography
\usepackage{enumitem}
\usepackage{times}
\usepackage{latexsym}
\usepackage{graphicx}
\usepackage{multicol}
\usepackage{multirow}
\usepackage{amsmath}
\usepackage{amssymb}
\usepackage{bbding}
\usepackage{subfigure}
\usepackage{colortbl}
\usepackage{pifont}
\usepackage{makecell}
\usepackage{mathrsfs}
\usepackage{bm}
\usepackage{bbm}
\usepackage{dsfont}
\usepackage{pgf}
\usepackage{fontawesome}
\usepackage{wrapfig}
\usepackage{tcolorbox}
\usepackage{array}
\usepackage{longtable} 
\usepackage{caption}
\definecolor{darkgray}{gray}{0.3}
\definecolor{lightblue}{RGB}{115, 192, 222}

\newtcolorbox{promptbox}[2][Prompt]{
colback=black!5!white,
arc=4pt, 
boxrule=0.5pt,
fonttitle=\bfseries,
title=#1, 
before upper={\small}, fontupper=\fontfamily{ptm}\selectfont,
colframe=#2,
}

\NewDocumentCommand\emojidifficulty{}{
$\vcenter{\hbox{\includegraphics[height=1.0em]{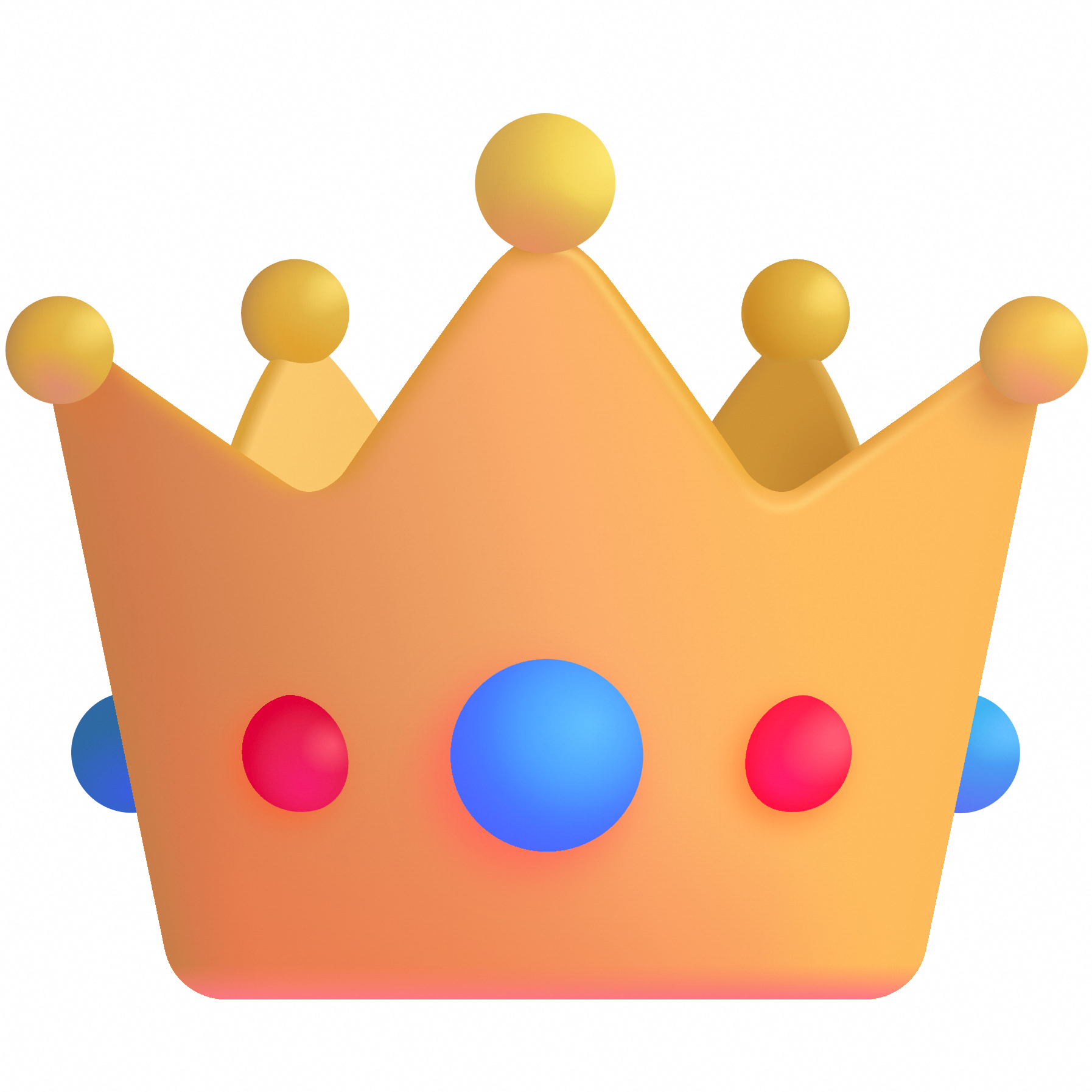}}}$
}

\newcommand{\baseline}{ControlledCoder}
\newcommand{\benchmark}{IFEvalCode}
\newcommand{\instruct}{IFEvalCode-Instruct}

\title{\benchmark{}: Controlled Code Generation}

\author{
  Jian Yang\textsuperscript{\rm 1}, 
  Wei Zhang\textsuperscript{\rm 1\thanks{Corresponding Author}},
  {\bf Shukai Liu}\textsuperscript{\rm 1},
  {\bf Linzheng Chai}\textsuperscript{\rm 1},
  {\bf Yingshui Tan}\textsuperscript{\rm 2}, 
  {\bf Jiaheng Liu}\textsuperscript{\rm 1}, \\ 
  {\bf Ge Zhang}\textsuperscript{\rm 2},
  {\bf Wangchunshu Zhou}\textsuperscript{\rm 3},
  {\bf Guanglin Niu}\textsuperscript{\rm 1$^\text{*}$}, 
  {\bf Zhoujun Li\textsuperscript{\rm 1$^\text{*}$}}, 
  {\bf Aishan Liu}\textsuperscript{\rm 1},
  {\bf Binyuan Hui}, 
  {\bf Junyang Lin}  \\
 \textsuperscript{\rm 1}Beihang University,~\textsuperscript{\rm 2}{\rm{M-A-P}},~\textsuperscript{\rm 3}{\rm{OPPO}}  \\
  \texttt{jiayang@buaa.edu.cn} \\
  \url{https://ifevalcode.github.io/}
}

\begin{document}

\maketitle

\begin{abstract}
Code large language models (Code LLMs) have achieved significant advancements in various code-related tasks, particularly in code generation, where the code LLMs produce the target code from natural language descriptions. However, in realistic scenarios, users often expect the returned code to strictly follow the given detailed requirements in many aspects (e.g. the style of code, the number of code lines, or the number of lines), instead of only requiring the correctness of the generated code. Controlled code generation means that the generated response from code LLMs should adhere to specific human guidelines or standards, whereas the LLM should have a strong instruction-following capability in the field of the code. In this paper, we propose forward constraints generation and backward constraints generation for controlled code generation to enhance the capability of LLM in following human instructions. Then, we build a multilingual benchmark \benchmark{} to evaluate the code instruction-following capability of the LLMs. \benchmark{} consists of 1.6K samples (Python, Java, Javascript, Typescript, Shell, Cpp, Php, and C-sharp) and each test sample contains the Chinese and English query. Different from the existing code benchmarks, we separately design the test function to verify the correctness of the code (Corr.) and whether the generated code follows the human instruction (Instr.). Extensive experimental results on \benchmark{} of 40+ LLMs emphasize that closed-source LLMs still dominate in controllable code generation compared to open-source LLMs, and the ability of LLMs to generate controllable code is far behind its ability to generate correct code. 

\end{abstract}

\section{Introduction}
\label{sec:introduction}
Large Language Models (LLMs) have made significant strides in code-related tasks, particularly in producing target code snippets from the given human description. It has captured considerable attention from academia and industry, owing to its practical relevance in software development. The current top-tier closed-source LLMs, such as GPT4.5~\citep{gpt45} and Claude3.7~\citep{claude37}, can correctly generate the file-level code snippet most of the time. Besides, a series of open-source code LLMs, such as StarCoder1/2~\citep{starcoder,starcoder2}, Deepseek-Coder~\citep{deepseek_coder}, OpenCoder~\citep{opencoder}, and Qwen2.5-Coder~\citep{qwen_coder}, has also become a force that cannot be ignored by industry and academia of code intelligence.

Most existing code benchmarks~\citep{mceval}, such as HumanEval~\citep{codex}, BigCodeBench~\citep{bigcodebench}, and MBPP~\citep{mbpp} are designed to evaluate the correctness of generated code by code execution with the unit tests. Further, some benchmarks, such as Chatbot Arena~\citep{llm_as_a_judge} and CodeArena~\citep{codearena} are proposed to evaluate the alignment between the model-generated response and human preference using the LLM-as-a-Judge. For the general domain, IFEval~\citep{ifeval} is proposed to evaluate the proficiency of LLM in controlled text generation, where the instructions are amenable to objective verification of compliance. In the field of code, in addition to requiring the generation of correct code, users often also require the generation of code meeting various objective requirements, such as code style, variable naming, specific algorithms, time complexity, etc. \textit{Therefore, we try to explore the proficiency of LLMs in controlled code generation by designing a framework to enhance and evaluate the capabilities in code instruction following.}

%%%%%%%%%%%%%%%%%%%%%%%%%%%%%%%%%%%%%%%%%%%%%%%%%%%%%%%%%%%
\begin{figure}[t]
\centering
\includegraphics[width=1\linewidth]{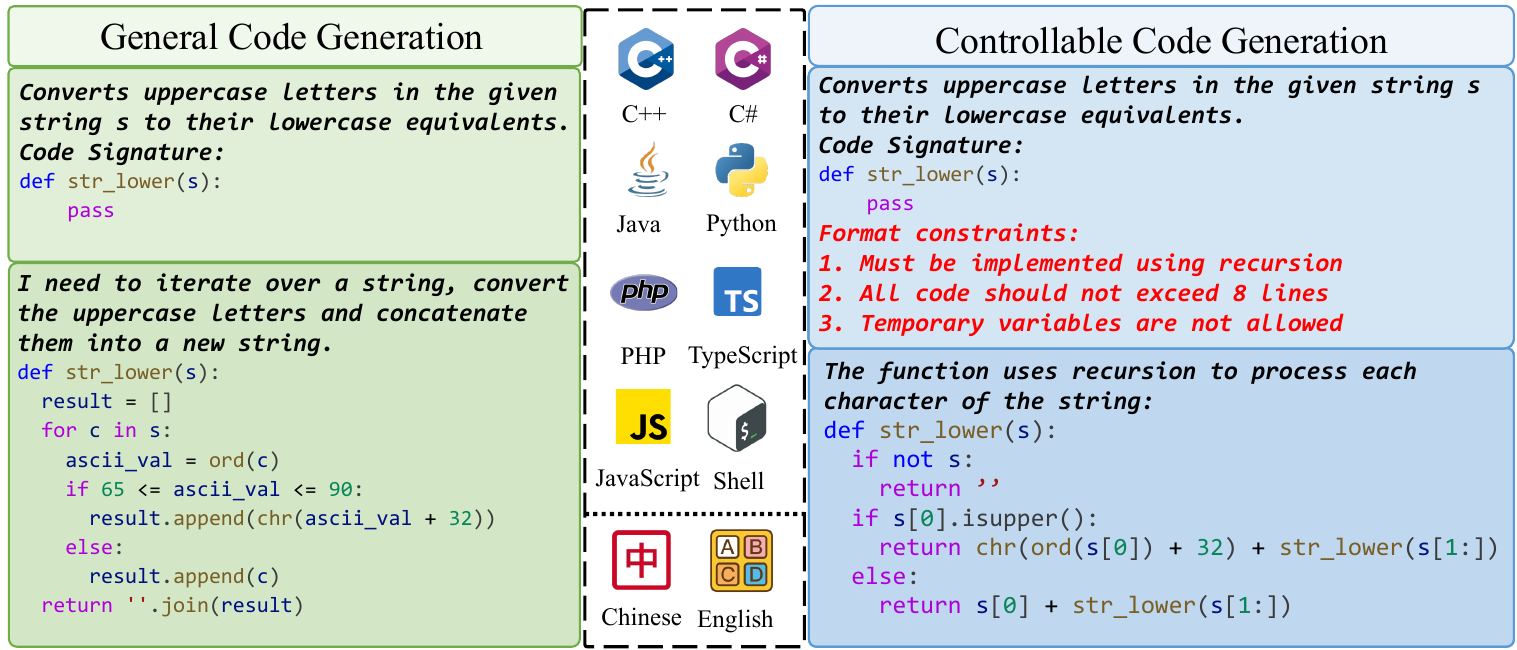}
\caption{Comparision between general code generation and Controllable code generation for 8 programming language and 2 natural language.}
\vspace{-15pt}
\label{fig:intro}
\end{figure}
%%%%%%%%%%%%%%%%%%%%%%%%%%%%%%%%%%%%%%%%%%%%%%%%%%%%%%%%%%%
To evaluate the capability of generating the controlled code, we construct a multilingual benchmark using human annotators, \benchmark{}, comprising nearly 1.6K samples across $8$ programming languages (Python, Java, CPP, C-Sharp, TypeScript, JavaScript, PHP, Shell), with each sample including both Chinese and English queries. Unlike existing benchmarks, \benchmark{} uniquely assesses two critical aspects: the correctness (code correctness) of the generated code and its adherence to human instructions (code controllability). 
In this paper, we introduce a novel framework combining forward constraints generation and backward generation for controlled code generation, aimed at enhancing the ability of LLM to follow human instructions in controlled code generation. 

The contributions are summarized as follows:
\begin{itemize}
    \item We introduce \benchmark{}, a multilingual benchmark comprising 1,620 samples across $8$ programming languages (Python, Java, JavaScript, TypeScript, Shell, C++, and C\#) with each sample featuring paired Chinese and English queries to facilitate comprehensive evaluation in controlled code generation.
    \item We integrate forward constraints generation and backward constraints generation to generate the instruction corpus \instruct{}, designed to enhance the capability of LLMs in adhering to human instructions during code generation. By jointly training to improve the controllability, accuracy, and alignment of generated code with specific user requirements, thereby advancing the performance of LLMs in structured and instruction-driven coding tasks.
    \item  We perform a comprehensive evaluation and analysis of existing 40+ LLMs on \benchmark{}, and several insightful findings are as follows: (1)
    Controlled code generation is challenging for LLMs. LLMs still have huge room for improvement and a long journey ahead to reach their full potential.
    (2) Closed-source models (e.g. Claude3.7) and LLMs with large sizes still significantly outperform open-source LLMs and LLMs with small sizes in controllable code generation. (3) The ability to generate specified code is significantly weaker than the ability to generate correct code.
\end{itemize}

\section{Multilingual Controlled Code Generation}
\subsection{Task Definition}
Given the $k$-th programming language $L_{k} \in \{L_{k}\}_{k=1}^{K}$ (in this work, $K=8$). We provide the problem description $q^{L_{k}}$ and instruction constraints $c^{L_{k}}=\{c^{L_{k}}_{1},\dots,c^{L_{k}}_{m}\}$ as input to the code LLMs $\mathcal{M}$ (Note: the constraints for code generation are language-specific). 
The LLMs generate the corresponding code $a^{L_{k}}$ by sampling from the code generation distribution $P(a^{L_{k}} | q^{L_{k}}, c^{L_{k}}; \mathcal{M})$. The generated code is then executed with the provided test cases $u_{L^{k}}$ to verify correctness. Specifically, we design another check function $v^{L_{k}}$ for each query to determine whether the generated code is controllable, meeting human instruction constraints. This process ensures that the generated code $a^{L_{k}}$ functionally aligns with the problem requirements and :
\begin{MiddleEquation}
\begin{align}
    r^{L_{k}}_{u} = \mathbb{I}(P(a^{L_{k}}|q^{L_{k}};\mathcal{M}); u^{L_{k}}); \ \ \ \ \ \
    r^{L_{k}}_{v} = \mathbb{I}(P(a^{L_{k}}|q^{L_{k}};\mathcal{M}); v^{L_{k}});
    \label{eval_code_generation}
\end{align}
\end{MiddleEquation}where $\mathbb{I}(\cdot)$ is the code execution in a sandbox environment with the given test cases $u^{L_{k}}$ or the check function $v^{L_{k}}$. when the generated code $a^{L_{k}}$ passes the unit tests $u^{L_{k}}$, the evaluation result $r^{L_{k}}_{u}=1$, else $r^{L_{k}}_{u}=0$. For controlled code generation, 
the check function $v^{L_{k}}$ is used to check whether the generated code satisfies the human instruction. $u^{L_{k}}$ and $v^{L_{k}}$ separately denote the scores of code correctness and code controllability.

\subsection{Overview of \benchmark{}}
%%%%%%%%%%%%%%%%%%%%%%%%%%%%%%%%%%%%%%%%%%%%%%%%%%%%%%%%%%%
%to do
\begin{wraptable}[14]{r}{0.3\textwidth}
\caption{Benchmark statistics. }
\resizebox{0.3\textwidth}{!}{
\begin{tabular}{lr}
\hline
\toprule
\textbf{Statistics}            & \textbf{Number (Zh/En)}         \\
\midrule
\textbf{Problems}              & $810/810$                \\
Python                         & $103/103$                \\
Java                           & $102/102$                \\
Cpp                            & $100/100$                \\
Typescript                     & $100/100$                \\
Javascript                     & $100/100$                \\
Php                            & $100/100$                \\
Shell                          & $100/100$                \\
C-Sharp                          & $100/100$                \\
\midrule
\textbf{\#Instruction Constraints } &           \\
- min/max/avg length   &2/15/5    \\      
\midrule
\textbf{Length}                                         \\
Chinese Question \\
min/max/avg   & $142$/$1,373$/$358$ tokens          \\
English Question  \\
min/max/avg   & $135$/$1,304$/$340$ tokens          \\
\bottomrule
\hline
\end{tabular}}
\label{tab:detail_data}
\end{wraptable}
%%%%%%%%%%%%%%%%%%%%%%%%%%%%%%%%%%%%%%%%%%%%%%%%%%%%%%%%%%%%

In \autoref{tab:detail_data}, \benchmark{} consists of 1.6K unique problems for 8 programming languages. The benchmarks can be used for bilingual natural languages, including Chinese and English (1.6K for English and 1.6K translated queries for Chinese). Each sample in \benchmark{} includes (\textit{en\_question, zh\_question, check\_instruction, check\_correctness}), where `\textit{check\_instruction}` is used to check the code correctness and `\textit{check\_correctness}` judges whether the generated code follows the human instruction. We calculate the length of the question using the Qwen2.5-Coder tokenizer \citep{qwen_coder} and count the number of instruction constraints for each sample. Each question contains 3 constraints and 100 tokens on average, where each prompt contains a problem description and the corresponding function or class signature.

In \autoref{tab:compare_bench}, we compare \benchmark{} with other code evaluation benchmarks in 6 aspects. Considering the evaluation cost and effectiveness, \benchmark{} has 1620 samples for 8 programming languages and 2 human languages. Specifically, we introduce the two parts of unit tests to check the code correctness and whether the code follows the human instruction.

%%%%%%%%%%%%%%%%%%%%%%%%%%%%%%%%%%%%%%%%%%%%%%%%%%%%%%%%%%%%%%%%%%%%%%%%%%%%%%%%%%%%%%%%%%%%%%%%%%%%%
\begin{table*}[h]
    \centering
    \caption{Comparison between \benchmark{} and other code benchmarks. Different from the existing benchmarks, \benchmark{} effectively provides a new comprehensive view of controlled code generation by adding the `\textit{check\_instruction}` part into the test samples.}
\resizebox{1.0 \textwidth}{!}{
    \begin{tabular}{l|ccclcc}
    \toprule
\bf Benchmark & \bf \#Languages & \bf \#Domains & \bf \makecell[c]{Size \\ (Zh/En)} & \bf \makecell[c]{Difficulty} & \bf \makecell[c]{Human \\ Check} & \bf \makecell[c]{Check Code \\ Controllability} \\
\midrule 
HumanEval~\citep{codex} & Python  & 3 & 164 (En) & \emojidifficulty{} & \textcolor{green}{\ding{51}} & \textcolor{red}{\ding{55}} \\
MBPP~\citep{mbpp} & Python  & Algorithm & 974 (En) & \emojidifficulty{} & \textcolor{green}{\ding{51}} & \textcolor{red}{\ding{55}} \\
BigCodeBench~\citep{bigcodebench} & Python & 7 & 2000+ (En) & \emojidifficulty{}\emojidifficulty{} & \textcolor{red}{\ding{55}} & \textcolor{red}{\ding{55}}\\
CodeArena~\citep{codearena} & 44 & 7 & 397+ (En) & \emojidifficulty{}\emojidifficulty{} & \textcolor{red}{\ding{55}} & \textcolor{red}{\ding{55}}\\
FullStackBench~\citep{fullstackbench} & 16 & 11 & 3374 (En) & \emojidifficulty{}\emojidifficulty{} & \textcolor{green}{\ding{51}} & \textcolor{red}{\ding{55}}\\
LiveCodeBench~\citep{livecodebench} & Python & 3 & 450+ (En) & \emojidifficulty{}\emojidifficulty{}\emojidifficulty{} & \textcolor{red}{\ding{55}} & \textcolor{red}{\ding{55}}\\
MultiPL-E~\citep{multipl_e} & 19 & 3 & 12,667 (En) & \emojidifficulty{}\emojidifficulty{} & \textcolor{red}{\ding{55}} & \textcolor{red}{\ding{55}}\\
McEval~\citep{mceval} & 20 & 10+ & 16K (En) & \emojidifficulty{}\emojidifficulty{}\emojidifficulty{} & \textcolor{green}{\ding{51}} & \textcolor{red}{\ding{55}}\\
MdEval~\citep{mdeval} & 40 & 10+ & 39K (En) & \emojidifficulty{}\emojidifficulty{}\emojidifficulty{} & \textcolor{green}{\ding{51}} & \textcolor{red}{\ding{55}}\\
DS-1000~\citep{ds1000} & Python & Data Size  & 1000 (En)  & \emojidifficulty{}\emojidifficulty{} & \textcolor{green}{\ding{51}} & \textcolor{red}{\ding{55}} \\
\midrule
\benchmark{}~(Ours) & 8 & 3 & 810/810 & \emojidifficulty{}\emojidifficulty{}\emojidifficulty{} & \textcolor{green}{\ding{51}} & \textcolor{green}{\ding{51}}\\ 
\bottomrule
\end{tabular}
}
\label{tab:compare_bench}
\end{table*}
%%%%%%%%%%%%%%%%%%%%%%%%%%%%%%%%%%%%%%%%%%%%%%%%%%%%%%%%%%%%%%%%%%%%%%%%%%%%%%%%%%%%%%%%%%%%%%%%%%%%%

%to do
\begin{figure*}[htb]
	\centering
	\subfigure[Domain types.]{\includegraphics[width=0.35\textwidth]{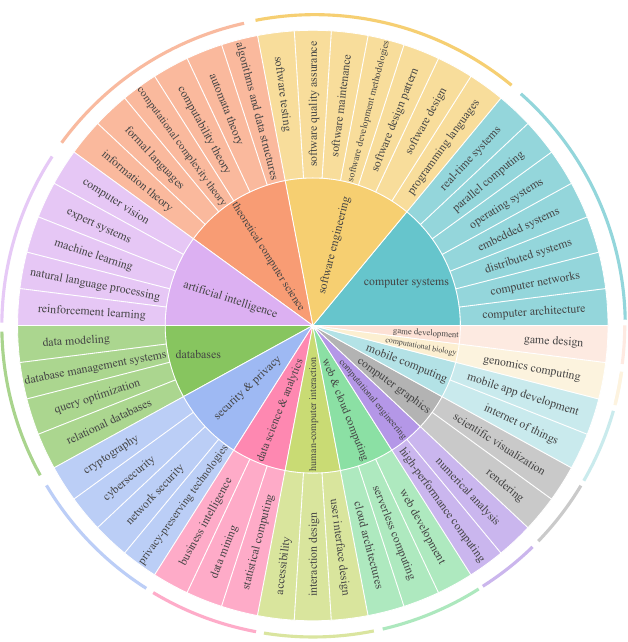}
    \label{fig:domain_types}}
    \hspace{0.1\textwidth} % 调整子图间距
	\subfigure[Types of instruction constraints.]{\includegraphics[width=0.35\textwidth]
    {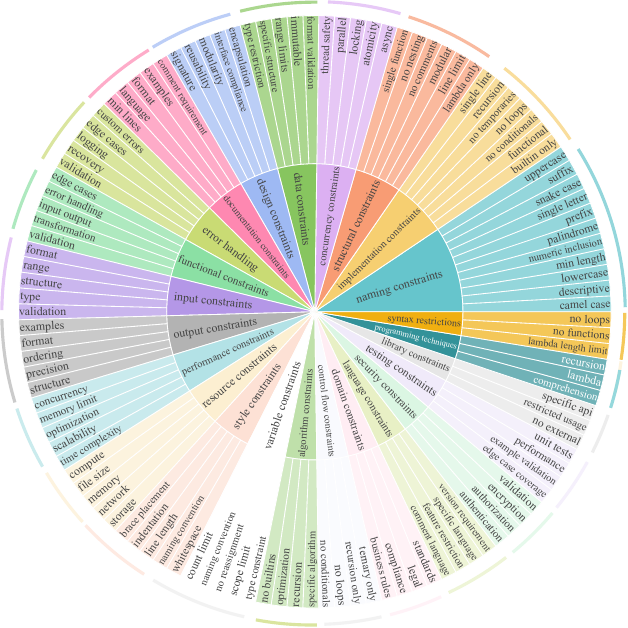}
    \label{fig:instruction_types}}
	\caption{Figure \autoref{fig:domain_types} shows domain types and Figure \autoref{fig:instruction_types} lists instruction types.}
	\label{fig:error_types}
\end{figure*} 

\subsection{\benchmark{} Construction \& Quality Control}
\begin{figure*}[h]
\begin{center}
    \includegraphics[width=1.0\textwidth]{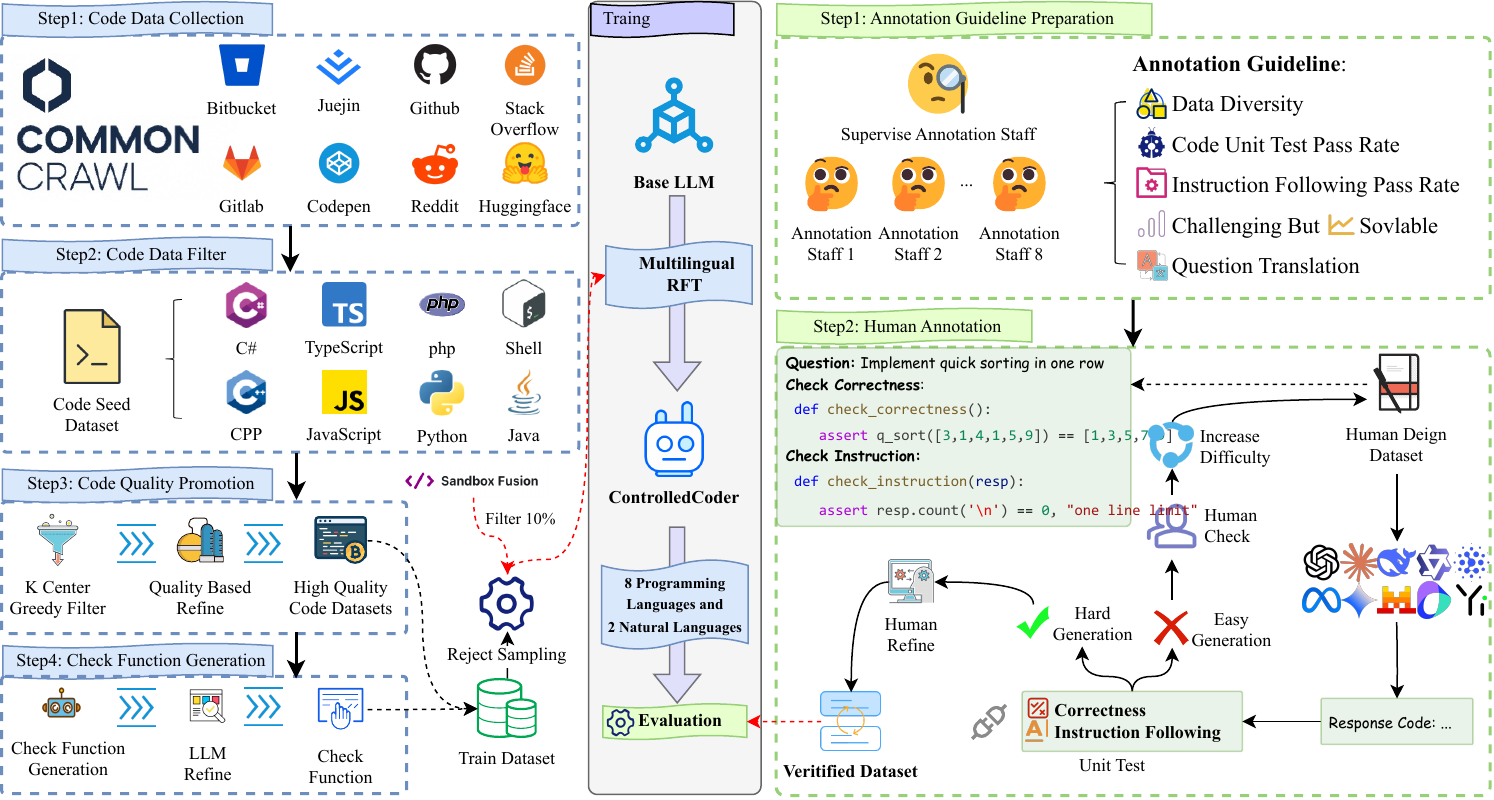}
    \vspace{-5pt}
    \caption{Construction of the benchmark \benchmark{} and instruction corpus \instruct{}.}
    \label{fig:framework}
    \vspace{-15pt}
\end{center}
\end{figure*}

To create \benchmark{} for controlled code generation, we propose a systematic human annotation and check procedure, which is guided by defined guidelines to guarantee accuracy and consistency. \autoref{fig:framework} illustrates the overall process of dataset construction. We begin by collecting code snippets and code-related documents from websites and recruit 6 computer science graduates as annotators. The annotators must follow the provided guidelines (1) ensure the diversity of the questions. (2) the question and instruction constraints should be challenging for existing LLMs. (3) translate the English question into the Chinese question. To increase the difficulty of \benchmark{}, the annotator filter out the questions, which half LLMs (GPT4o, DeepSeek-V3, Claude3.7, etc) can correctly answer.

\begin{figure*}[t]
\begin{center}
    \includegraphics[width=1.0\textwidth]{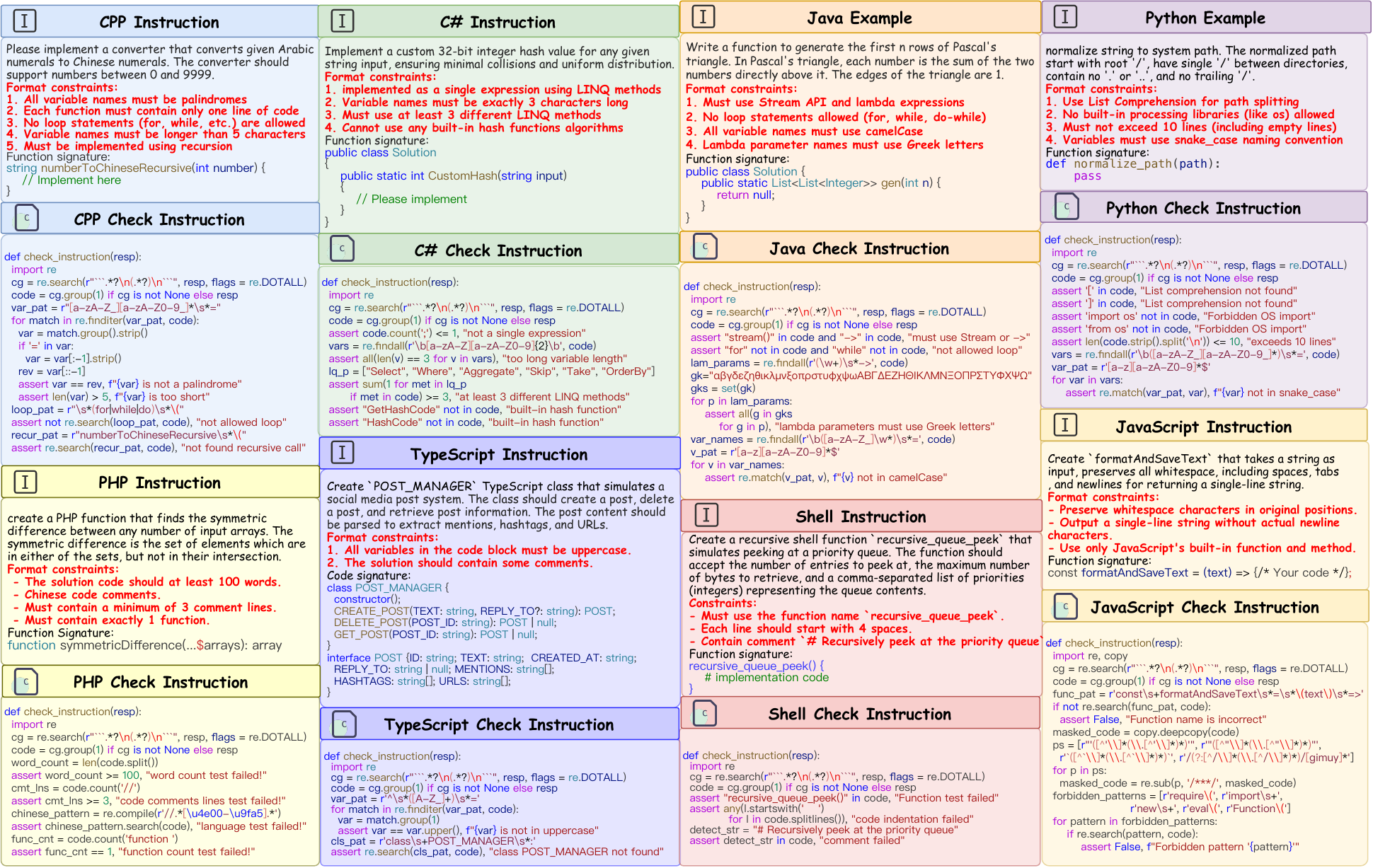}
    \vspace{-5pt}
    \caption{Examples of the verifiable instructions with `\textit{check\_instruction}' in  \benchmark{}. }
    \label{fig:examples}
    \vspace{-10pt}
\end{center}
\end{figure*}

\subsection{Reject-sampling Fine-tuning for Controlled Code Generation}
\paragraph{Creating Queries and Unit Tests with Constraints}To create the instruction corpora, we collect the code-related websites and remove the HTML tags to obtain the code snippet with the corresponding natural language description as the seed data. Then, we select the select seed data from 8 languages and force the code LLM (Qwen2.5-Coder 32B) to generate a challenging question $q^{L_{k}}$ of programming language $L_{k}$. To verify the correctness of the response, we prompt the LLM to generate the test cases $u_{k}$ to create `\textit{check\_correctness}'. To independently check whether the model-generated response satisfies the constraints in the question, we assume that the response is correct and only check whether the generation code follows the user instructions. Therefore, for each query $q^{L_{k}}$ with constraints $c^{L_{k}}$, we create the unit tests $r_{u}^{L_{k}}$ to check the correctness of the response and $r_{u}^{L_{k}}$ to check the whether the generated code is under control. 

\paragraph{Forward Constraints Generation} Given the recalled code-related documents from Common Crawl, we adopt Qwen2.5-Coder-32B to create new questions by drawing inspiration from the code-related documents for a general question. To effectively inject the additional instruction constraints into the question, we select the constraint types from the constraints types pool $C=\{c_{i=1}\}_{i=1}^{m}$ and force the model to generate a new question $q^{L_{k}}$ with constraints. After generating the question $q^{L_{k}}$ with constraints $c^{L_{k}}$, we force the LLM to generate $r^{L_{k}}_{u}$ and $r^{L_{k}}_{v}$ and the only keep the samples passing both unit tests. The generated corpus is denoted as $D_{s_1}$

\paragraph{Backward Constraints Generation} Since the forward constraints generation must first pre-define the constraint types and then generate the corresponding question, we introduce the backward constraints generation to synthesize the question with random instruction constraints. First, we prompt the LLM to generate the question $q^{L_{k}}$ and answer $a^{L_{k}}$. Then, we prompt the LLM to summarize which constraints the generated code complies with and obtain othe constraints $c^{L_{k}}$. Finally, we generate the unit tests $r^{L_{k}}_{u}$ and $r^{L_{k}}_{v}$, which are further used to verify the response and the created instruction corpus from backward constraints generation is denoted as $D_{s_2}^{L_{k}}$.

\paragraph{Joint Multilingual Rejecting Fine-tuning for Controlled Code Generation} 
Given the large-scale instruction corpus $D^{L_{k}} = D_{s_1}^{L_{k}}\bigcup D_{s_{2}}^{L_{k}}$ of different programming languages, the LLM is jointly optimized for response generation and unit test generation of different programming languages:
\begin{MiddleEquation}
\begin{align}
    \mathcal{L}_{all} = -\sum_{k=1}^{K} \mathbb{E}_{D^{L_{k}} }  \log P(a^{L_{k}}|q^{L_{k}},c^{L_{k}})  -\sum_{k=1}^{K} \mathbb{E}_{ D^{L_{k}} }  \log P(u^{L_{k}}_{v},u^{L_{k}}_{r}|q^{L_{k}},c^{L_{k}}) 
    \label{equ:multilingual_loss}
\end{align}
\end{MiddleEquation}where the responses from $D^{L_{k}}$ pass the unit tests of correctness and instruction-following. We can directly optimize the LLM by generating answers and unit tests.

\section{Experiments}
\subsection{Experiment Setup}
\label{sec:experimental_setup}

\paragraph{Evaluation Metric}
To check whether the model-generated response is correct (`Corr.': Correctness Pass@$1$) and follows the user instructions (`Instr.': Instruction-following Pass@$1$), we evaluate LLMs by executing the generated code with unit tests (pass rate for just one-time generation) and separately report `Corr.' and `Instr.' Pass@$1$ score in tables.

\paragraph{Code LLMs}
We report 40+ popular LLMs for \benchmark{}, including o1-mini~\citep{o1_mini}, GPT4o~\citep{gpt4}, Claude-3.5/3.7~\citep{claude,claude37}, Deepseek-v3/R1~\citep{deepseekv3,deepseek_r1} llama3~\citep{llama3}, qwen3~\citep{qwen}, and code-specific models like Qwen2.5-Coder~\citep{qwencoder}, DeepSeekCoder~\citep{deepseek_coder}, CodeLlama~\citep{code_llama}, and Codegemma~\citep{codegemma}. Additionally, we fine-tune Qwen2.5-Coder-32B as our baseline \baseline{}. 

\paragraph{\baseline{} Training Setup}
Our constructed code generation dataset \instruct{} is for training \baseline{}, ensuring fundamental instruction-following capabilities for code-related tasks. \baseline{}, built on Qwen2.5-Coder-32B, is trained for 3 epochs using a cosine scheduler with an initial learning rate of \(5 \times 10^{-5}\) with a 3\% warmup ratio. We employ AdamW~\citep{adamw} as the optimizer, with a batch size of 1024 and a maximum sequence length of 2048. (Details can be found in the Appendix).

\input{tables/table_en}

\input{tables/table_zh}

\subsection{Main Results}

\paragraph{Results of English queries in \benchmark{}} Table~\ref{tab:ifevalcode_en} presents evaluation results of ``Correctness'' (Corr.) and ``Instruction Compliance'' (Instr.) for 40+ LLMs across $8$ programming languages (Python, Java, C++, C\#, TypeScript, JavaScript, PHP, Shell). These models vary in size and origin (closed\-source/open\-source/mixed expert models) and are tested on English programming queries. Closed-source LLMs (such as Claude-3.7, GPT-4.1, and various mini versions) generally achieve overall correctness rates concentrated between 30\% and 40\%. Instruction compliance rates typically hover around 20\% to 25\%. Open-source LLMs with 0.5B to 3B parameters generally demonstrate limited performance, with average correctness rates mostly between 6\% and 18\%. The top performer in this range, OpenCoder (1.5B), only reaches 17.9\%. This indicates that small-scale models still struggle to achieve satisfactory results in cross-language, multi-task coding scenarios. Open-source models in the 6B to 14B parameter range show significant improvement. The 7B Qwen2.5-Coder, 8B Qwen3-think, and 14B Qwen3 models all surpass average correctness rates of 25\% to 30\%. Notably, the 14B Qwen3-think achieves a multilingual overall correctness rate of 29.6\%. Among models with 32B parameters or more / MoE architectures, Deepseek-R1 (37/637B) performs the best, reaching an average correctness rate as high as 43.1\%. It sets the highest group scores across multiple languages: Python (58.1\%), Java (37.3\%), C++ (38.0\%), TypeScript (46.0\%), JavaScript (55.0\%), PHP (61.0\%), and Shell (45.0\%). Its instruction compliance rate also improves to around 19\% to 20\%. Overall, model scale and architectural complexity (such as MoE expert models) significantly boost programming task capabilities. At the same time, whether closed-source or open-source, the current instruction compliance rates remain noticeably lower than correctness rates. This highlights the need for further optimization in instruction tuning and code execution verification in the future.

\paragraph{Results of Chinese queries in \benchmark{}} 
Table~\ref{tab:ifevalcode_zh} presents the evaluation results of various large models on Chinese programming queries (covering eight languages: Python, Java, C++, C\#, TypeScript, JavaScript, PHP, and Shell) in terms of "Correctness" (Corr.) and "Instruction Compliance" (Instr.). The best performers within each size range are highlighted in bold. Closed-source models (Claude-3.x, GPT-4.0/4.1, and their mini versions) have an overall average correctness between 27\% and 38\%, with instruction compliance around 20\% to 25\%. Among them, GPT-4.1 leads the closed-source group in both correctness (36.4\%) and instruction compliance (24.7\%). Although o1-mini and o3-mini show higher correctness rates (39.3\% and 37.8\%, respectively), their instruction compliance rates are relatively lower (23.5\% and 24.1\%). Open-source models in the 0.5B–3B parameter range show limited performance, with average correctness mostly between 5\% and 18\%. OpenCoder (1.5B) at 18.0\% and Qwen2.5-Coder (3B) also at 18.0\% lead this group, indicating that small-scale models still need improvement in multilingual coding tasks. In the 6B–14B parameter range, open-source models begin to show significant improvement—Yi-Coder (9B) achieves an average correctness of 26.9\%, while Qwen2.5-Coder (14B) ranks first in this group with a correctness of 29.9\% and instruction compliance of 19.6\%. Qwen3 (8B) also surpasses a 23\% correctness rate and obtains 32\% instruction compliance on Python and TypeScript. Among large models with 32B+ parameters and MoE (Mixture of Experts) architectures, Deepseek-R1 (37/671B) demonstrates the most outstanding performance, with an average correctness of 43.1\% and the highest scores across multiple sub-tasks including Python, Java, C++, TypeScript, JavaScript, PHP, and Shell. Although its instruction compliance (18.8\%) is not the highest, it remains relatively high. Deepseek-V3 follows closely (34.6\%), with Qwen2.5-Coder 32B (29.6\%) and Llama3.1-70B (25.9\%) ranking next. Overall, model scale and architecture (especially MoE expert models) have a significant positive impact on Chinese programming capabilities. Meanwhile, instruction compliance rates are generally lower than correctness rates across all models, highlighting the need for further optimization in instruction tuning and multilingual semantic understanding in the future.

\section{Further Analysis}
\paragraph{Ablation Study.}
%%%%%%%%%%%%%%%%%%%%%%%%%%%%%%%%%%%%%%%%%%%%%%%%%%%%%%%%%%%%%%%%%%%%%%%%%%%%
\begin{wraptable}{r}{0.35\textwidth}
\centering
\caption{Ablation study.}
\resizebox{0.35\textwidth}{!}{
\begin{tabular}{c|c|cc}
\toprule
ID & Method          &  Corr.$_{avg}$ & Instr.$_{avg}$  \\
\midrule
{\large{\ding{172}}}  &     Qwen2.5-Coder-32B       &     29.9  &  20.7 \\
{\large{\ding{173}}} & {\large{\ding{172}}} + $D_{s_{1}}$             &  34.5    &  26.4    \\
{\large{\ding{174}}} & {\large{\ding{173}}} + $D_{s_{2}}$   &  38.2     &   30.4      \\
{\large{\ding{175}}} & {\large{\ding{174}}} + Iter2                   &  40.3   &  34.2    \\
{\large{\ding{176}}} & {\large{\ding{175}}} + Iter3                   &  40.5   &  33.2    \\
\bottomrule
\end{tabular}
}
\label{fig:ablation_study}
\end{wraptable}
%%%%%%%%%%%%%%%%%%%%%%%%%%%%%%%%%%%%%%%%%%%%%%%%%%%%%%%%%%%%%%%%%%%%%%%%%%%%
In \autoref{fig:ablation_study}, we conduct the ablation study to explore the effect of each component in \instruct{}. \baseline{} is fine-tuned on the base Qwen2.5-Coder by combining the instruction corpus from the forward and backward constraints generation ( {\large{\ding{173}}} and  {\large{\ding{174}}} outperform the baseline Qwen2.5-Coder {\large{\ding{172}}}). The fine-tuned coder \baseline{} can iteratively update the instruction corpus \instruct{} using forward and backward constraints generation. We can see that joint multilingual rejecting fine-tuning can effectively bring improvement for controlled code generation.

% %%%%%%%%%%%%%%%%%%%%%%%%%%%%%%%%%%%%%%%%%%%%%%%%%%%%%%%%%%%
\begin{figure*}[h]
	\centering
	\subfigure[Distribution of constraints number]{\label{constraints_num_distribution}\includegraphics[width=0.4\textwidth]{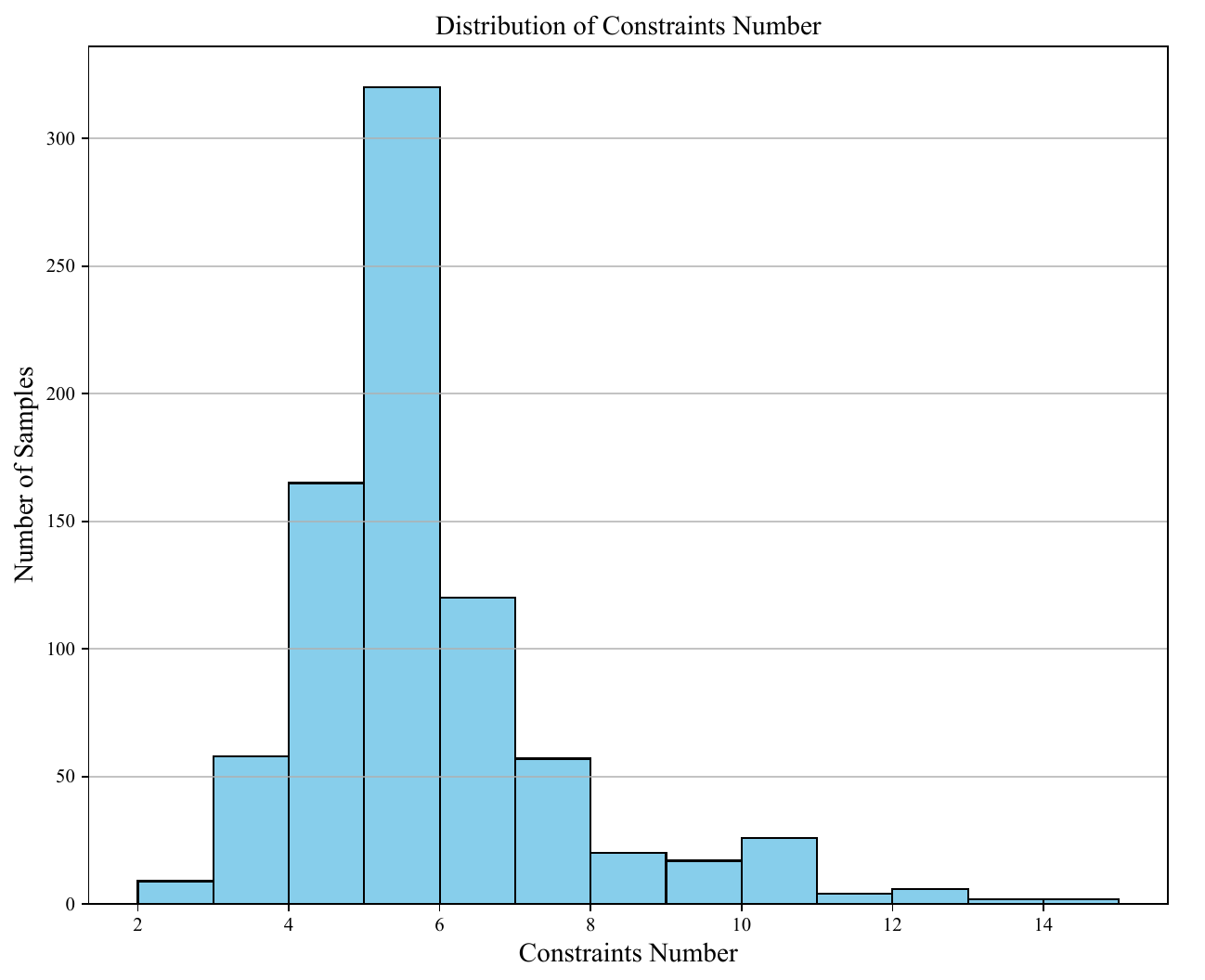}}
	\subfigure[Scores of different constraints number]{ \label{constraints_num}\includegraphics[width=0.4\textwidth]{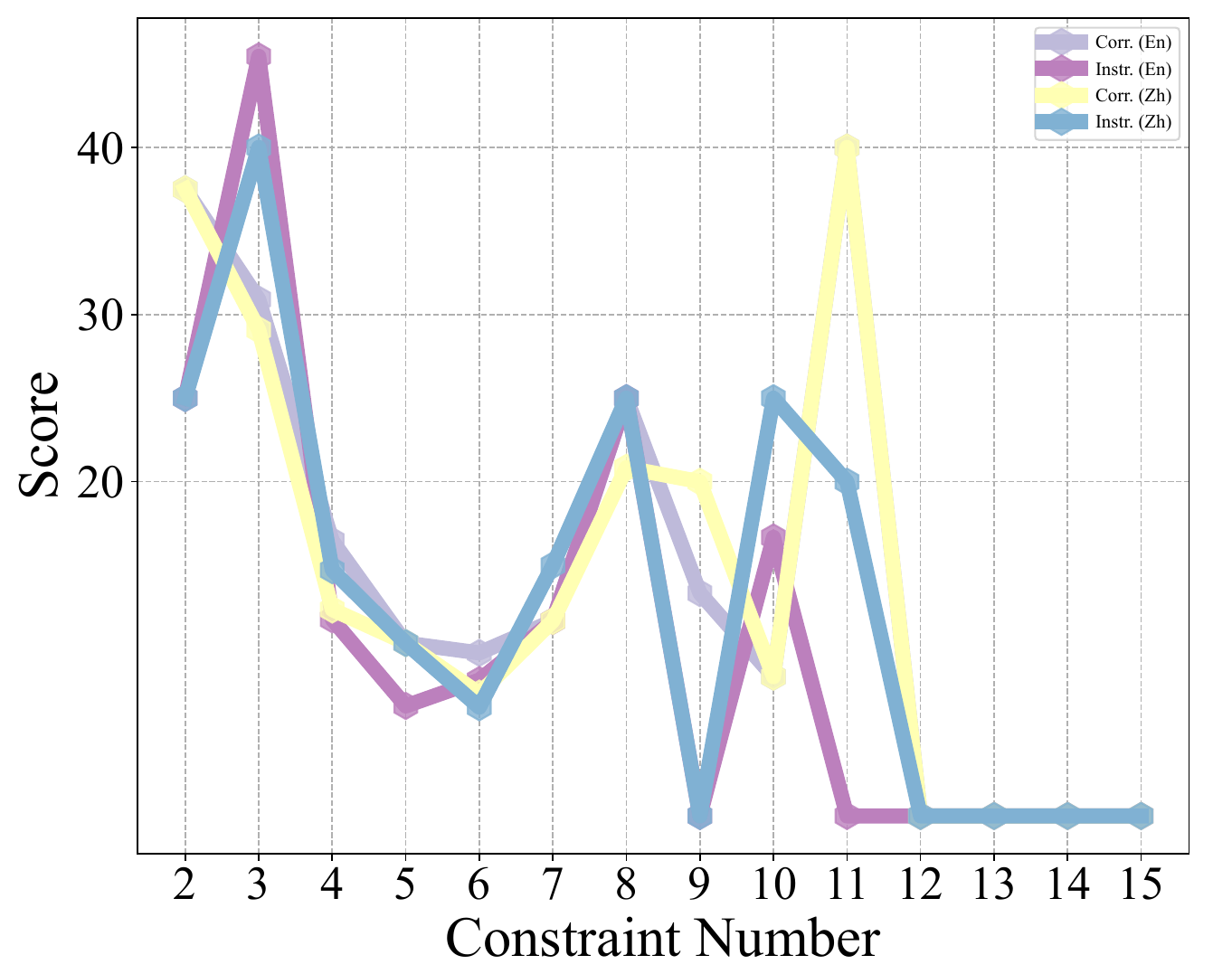}}
	\caption{Figure \autoref{constraints_num_distribution} and \autoref{constraints_num} plot the trends with the number of constraints increaing}
	\label{fig:repair_setting23}
\end{figure*} 
% %%%%%%%%%%%%%%%%%%%%%%%%%%%%%%%%%%%%%%%%%%%%%%%%%%%%%%%%%%%%

\begin{figure*}[h!]
\begin{center}
    \includegraphics[width=0.7\textwidth]{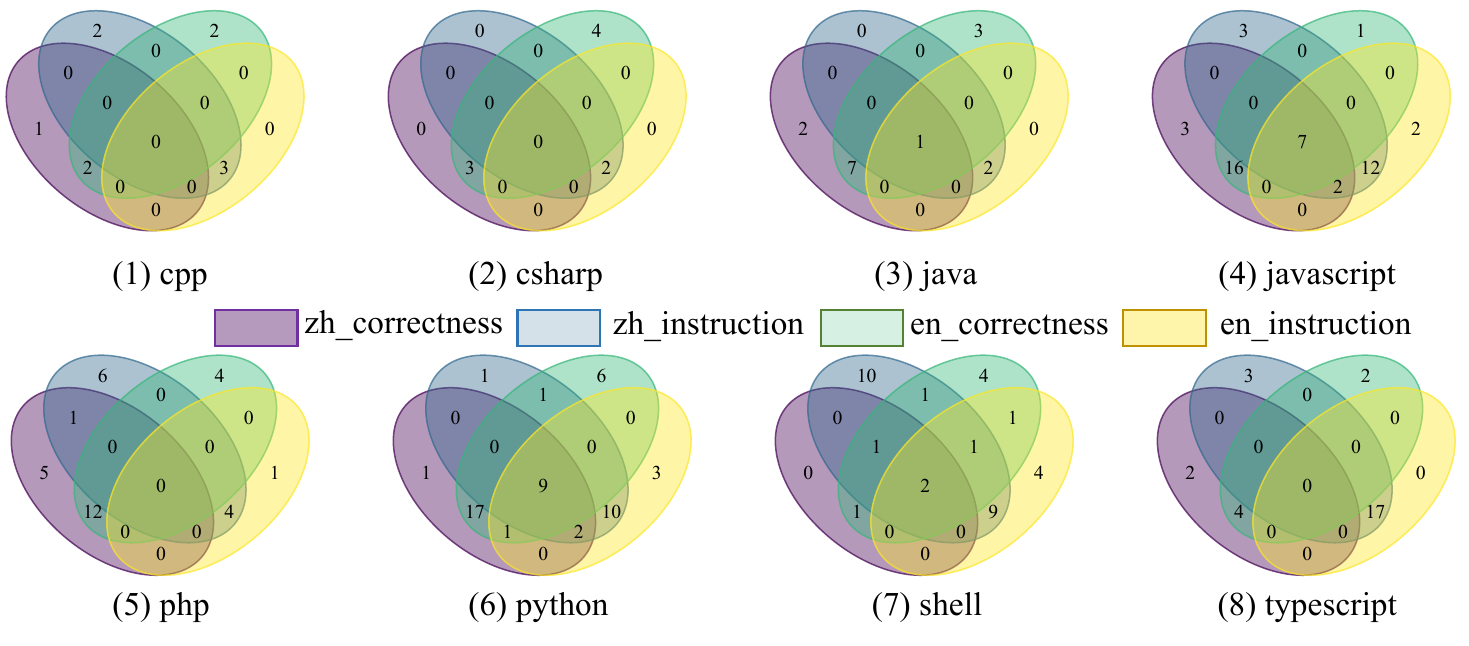}
    \caption{The Venn diagram of the numbers of the correct samples for each programming language. }
    \label{fig:venn}
    \vspace{-10pt}
\end{center}
\end{figure*}

\paragraph{Correlation between Correctness and Code Instruction Following}
\autoref{fig:venn} displays the correctness overlap between correctness score (Eh), instruction following score (Zh), correctness score (En), and instruction following score (Zn) of $8$ different programming languages. 

Figure~\ref{fig:venn} shows a grid of four-set Venn diagrams—one subplot per programming language—where each diagram partitions the model’s outputs into the following four evaluation sets: (1) Chinese‐prompt correctness, (2) Chinese‐prompt instruction compliance, (3) English‐prompt correctness, and (4) English‐prompt instruction compliance. By inspecting the overlaps, we can immediately see several consistent patterns across all eight languages: Correctness dominates. Both the Chinese‐correctness and English‐correctness sets are substantially larger than their corresponding instruction‐compliance sets, indicating that LLMs more often produce functionally correct code than code that also respects the extra constraints. Cross-language consistency in correctness. The overlap between Chinese‐correctness and English‐correctness is very large in every language, showing that whether the prompt is given in Chinese or English has little effect on a model’s ability to generate a passing solution. Instruction compliance is more fragile. The Chinese‐instruction and English‐instruction sets are much smaller, and their overlap is modest. In other words, even if a model is correct in both languages, it frequently fails to satisfy the same set of style or structural constraints when the prompt language switches. The smallest region is the four-way intersection. Only a small fraction of examples are simultaneously correct and compliant in both Chinese and English, demonstrating that achieving full controlled generation across languages remains the hardest challenge.

\section{Related Work}
\label{sec:related_work}

\paragraph{Code Large Language Model} 
The rapid progress of large language models\citep{gpt4,llama2,llama3,deepseekv3,deepseek_r1} has enabled complex code-related tasks. Early models (e.g. BERT~\citep{bert} and GPT~\citep{gpt}) trained on billions of code snippets, focused on code understanding and generation~\citep{codex, code_bert, bloom, AlphaCode,codet5,santacoder}. Recent advances in domain-specific pre-training and instruction fine-tuning~\citep{kun,mammoth2} have enhanced models like Qwen2.5-Coder~\citep{qwen_coder} and Deepseek-Coder~\citep{deepseek_coder}, achieving strong performance in code completion, synthesis, and repair.

\paragraph{Code Benchmarks} 
In the field of code, a rich tapestry of benchmarks~\citep{codex,mbpp,humaneval_x,codereval,arcade_nl2code,humaneval_xl,xcodeeval,BabelCode} has been developed to address the challenges of assessing code quality, functionality, and efficiency. Current benchmarks support the evaluation of various tasks, including code generation (e.g. NCB~\citep{naturalcodebench} and MultiPL-E~\citep{multipl_e}), code repair~ (e.g. DebugBench~\citep{debugbench} and MdEval~\citep{mdeval}), code competition (e.g. LCB~\citep{livecodebench} and CodeElo~\citep{codeelo}), software development (e.g. Aider and SWE-Bench~\citep{swe_bench}). But existing code benchmarks ignore the multilingual evaluation of controlled code generation for LLMs. IFEval~\citep{ifeval} emphasizes the importance of evaluating the proficiency of LLMs in instruction following, where IFEval contains a series of ``verifiable instructions''. 

\section{Conclusion}
\label{sec:conclusion}
% \vspace{-8pt}
In this paper, we first introduce \benchmark{}, a multilingual benchmark with 1.6K samples across 8 programming languages, designed to evaluate both code correctness and code controllability in LLM-generated code. To address these challenges, we propose a novel framework that combines forward constraints generation and backward generation, enhancing LLMs' ability to follow human instructions for controlled code generation. Our approach advances the reliability and precision of LLMs in adhering to user intent while producing functionally accurate code.

\bibliography{neurips}
\bibliographystyle{plain}
%%%%%%%%%%%%%%%%%%%%%%%%%%%%%%%%%%%%%%%%%%%%%%%%%%%%%%%%%%%
\clearpage
\input{appendix}
%\input{neurips_checklist_2025}
%%%%%%%%%%%%%%%%%%%%%%%%%%%%%%%%%%%%%%%%%%%%%%%%%%%%%%%%%%%

\end{document}

%% file: tables/table_en.tex
\begin{table*}[]
\caption{English evaluation of English queries in \benchmark{} of 8 programming languages. `Corr.' denotes the correctness of the model-generated response, and `Instr.' denotes the instruction-following accuracy of the response. The underlined values are the maximum in each column within each colored subgroup.}
\resizebox{1.0\textwidth}{!}{
\begin{tabular}{lc|cccccccccccccccccc}
\toprule
\multirow{2}{*}{Models}  & \multirow{2}{*}{Params} & \multicolumn{2}{c}{Python} & \multicolumn{2}{c}{Java} & \multicolumn{2}{c}{Cpp} & \multicolumn{2}{c}{C-sharp} & \multicolumn{2}{c}{Typescript} & \multicolumn{2}{c}{Javascript} & \multicolumn{2}{c}{Php} & \multicolumn{2}{c}{Shell} & \multicolumn{2}{c}{Avg.} \\ \cmidrule{3-20} 
 &    & Corr.       & Instr.       & Corr.      & Instr.      & Corr.      & Instr.     & Corr.        & Instr.       & Corr.         & Instr.         & Corr.         & Instr.         & Corr.      & Instr.     & Corr.       & Instr.      & Corr.      & Instr.      \\ \midrule
\multicolumn{20}{c}{\textit{Closed-source LLMs}}   \\ \midrule
\rowcolor{cyan!15} Claude-3.5-Sonnet  & \faLock{}   &  48.6 & 26.7 & 33.3 & 10.8 & 19.0 & 9.0 & 12.6 & 2.9 & 24.0 & 25.0 & 48.0 & 32.0 & 41.0 & 20.0 & 28.0 & 34.0 & 31.9 & 20.0 \\  
\rowcolor{cyan!15} Claude-3.7-Sonnet  & \faLock{}   &  47.6 & 31.4 & \underline{37.3} & 7.8 & \underline{29.0} & 6.0 & 16.5 & 3.9 & 29.0 & \underline{28.0} & 54.0 & 34.0 & \underline{59.0} & 27.0 & 36.0 & 43.0 & 38.5 & 22.6  \\  
\rowcolor{cyan!15} GPT-4o  & \faLock{}   &   45.7 & \underline{36.2} & 25.5 & 5.9 & 22.0 & 5.0 & 15.5 & 3.9 & 30.0 & 25.0 & 47.0 & 26.0 & 51.0 & 18.0 & 32.0 & 44.0 & 33.6 & 20.5 \\
\rowcolor{cyan!15} GPT-4o-mini  & \faLock{}   &  49.5 & 30.5 & 27.5 & 10.8 & 10.0 & 5.0 & 7.8 & \underline{5.8} & 29.0 & 25.0 & 47.0 & 27.0 & 42.0 & 22.0 & 23.0 & 44.0 & 29.5 & 21.2 \\
\rowcolor{cyan!15} GPT-4.1  & \faLock{}   &  50.5 & 31.4 & 36.3 & 12.7 & 27.0 & 6.0 & 1.0 & 3.9 & 30.0 & 27.0 & \underline{58.0} & 25.0 & 58.0 & 26.0 & \underline{40.0} & 52.0 & 36.8 & 23.0\\
\rowcolor{cyan!15} GPT-4.1-mini  & \faLock{}   &  46.7 & 34.3 & 35.3 & 7.8 & 23.0 & 4.0 & 0.0 & 2.9 & 27.0 & 23.0 & 57.0 & \underline{30.0} & 50.0 & 29.0 & 38.0 & \underline{53.0} & 34.6 & 23.0 \\
\rowcolor{cyan!15} grok-3  & \faLock{}   &  25.7 & 29.5 & 18.6 & \underline{13.7} & 2.0 & 6.0 & 0.0 & 3.9 & 4.0 & 19.0 & 44.0 & 21.0 & 43.0 & 24.0 & 16.0 & 33.0 & 19.1 & 18.8 \\
\rowcolor{cyan!15} grok-3-mini-fast  & \faLock{}   & 9.5 & 34.3 & 29.4 & 10.8 & 21.0 & 5.0 & 1.9 & 3.9 & 28.0 & 24.0 & 53.0 & 29.0 & 33.0 & 4.0 & 14.0 & 23.0 & 28.8 & 16.8 \\
\midrule
\multicolumn{20}{c}{\textit{0.5B+ Open-source LLMs}}     \\ \midrule
\rowcolor{olive!15} Deepseek-Coder & 1.3B &   24.8 & 17.1 & \underline{19.6} & 2.0 & 16.0 & 3.0 & 0.0 & 1.9 & 0.0 & 1.0 & 22.0 & 13.0 & 10.0 & 3.0 & 4.0 & 9.0 & 12.1 & 6.3            \\ 
\rowcolor{olive!15} Qwen2.5-Coder &   0.5B &  18.1 & 19.0 & 8.8 & 2.0 & 8.0 & 2.0 & 9.7 & 1.9 & 2.0 & 18.0 & 13.0 & 12.0 & 11.0 & 4.0 & 3.0 & 9.0 & 9.3 & 8.5  \\ 
\rowcolor{olive!15} Qwen2.5-Coder &   1.5B &  32.4 & 22.9 & 10.8 & 2.9 & 3.0 & 3.0 & 1.9 & 1.9 & 6.0 & 17.0 & 24.0 & 21.0 & 16.0 & 5.0 & 11.0 & 17.0 & 13.2 & 11.4 \\ 
\rowcolor{olive!15} Qwen2.5-Coder &   3B  &  35.2 & 24.8 & 11.8 & 2.0 & 11.0 & 5.0 & 11.7 & 1.9 & 0.0 & 18.0 & 26.0 & \underline{29.0} & 22.0 & 12.0 & \underline{15.0} & 23.0 & 16.7 & \underline{14.4}        \\ 
\rowcolor{olive!15} Granite-Coder &   3B &  25.7 & 21.0 & 10.8 & \underline{4.9} & 7.0 & 3.0 & \underline{12.6} & \underline{3.9} & 6.0 & 14.0 & 18.0 & 11.0 & 15.0 & 1.0 & 5.0 & 12.0 & 12.6 & 8.9 \\ 
\rowcolor{olive!15} OpenCoder &      1.5B &  \underline{37.1} & 23.8 & 16.7 & 3.9 & \underline{25.0} & 3.0 & 4.9 & 1.9 & 9.0 & 14.0 & 22.0 & 13.0 & 21.0 & 4.0 & 7.0 & 18.0 & \underline{17.9} & 10.2   \\ 
\rowcolor{olive!15} Yi-Coder &   1.5B &    32.4 & 21.0 & 13.7 & 2.0 & 13.0 & 4.0 & 11.7 & 2.9 & 8.0 & 17.0 & 26.0 & 21.0 & 16.0 & 4.0 & 8.0 & 14.0 & 16.2 & 10.7   \\ 
\rowcolor{olive!15} Qwen3 & 0.6B & 14.3 & 23.8 & 2.9 & 2.9 & 0.0 & 5.0 & 0.0 & 2.9 & 2.0 & 24.0 & 18.0 & 22.0 & 10.0 & 7.0 & 2.0 & 14.0 & 6.2 & 12.7            \\ 
\rowcolor{olive!15} Qwen3-think &   0.6B  & 13.3 & 15.2 & 2.0 & 2.0 & 1.0 & 5.0 & 0.0 & 1.9 & 2.0 & 6.0 & 8.0 & 3.0 & 13.0 & 0.0 & 0.0 & 6.0 & 4.9 & 4.9        \\ 
\rowcolor{olive!15} Qwen3 &   1.7B  & 23.8 & 26.7 & 8.8 & 3.9 & 3.0 & 4.0 & 3.9 & 1.9 & 12.0 & 22.0 & 22.0 & 16.0 & 19.0 & 9.0 & 7.0 & 23.0 & 12.5 & 13.3 \\ 
\rowcolor{olive!15} Qwen3-think &   1.7B  & 25.7 & 21.0 & 3.9 & 2.9 & 3.0 & 2.0 & 0.0 & 2.9 & 10.0 & 11.0 & 22.0 & 17.0 & \underline{28.0} & 2.0 & 7.0 & 13.0 & 12.5 & 9.0 \\        
\rowcolor{olive!15} Qwen3 &   4B  & 26.7 & \underline{28.6} & 7.8 & 5.9 & 7.0 & \underline{6.0} & 1.9 & 1.9 & \underline{16.0} & \underline{25.0} & \underline{29.0} & 28.0 & \underline{28.0} & \underline{13.0} & 7.0 & \underline{38.0} & 15.4 & 18 \\ 
\rowcolor{olive!15} Qwen3-think &   4B  & 34.3 & 21.9 & 12.7 & 4.9 & 6.0 & 4.0 & 1.0 & 3.9 & 12.0 & 12.0 & \underline{29.0} & 20.0 & 25.0 & 2.0 & 13.0 & 17.0 & 16.7 & 10.7   \\ \midrule
\multicolumn{20}{c}{\textit{6B+ Open-source LLMs}}       \\ \midrule
\rowcolor{magenta!15}  CodeLlama &   7B &       18.1 & 22.9 & 5.9 & 3.9 & 3.0 & \underline{4.0} & 2.9 & 2.9 & 7.0 & 19.0 & 20.0 & 22.0 & 13.0 & 9.0 & 10.0 & 21.0 & 10.0 & 13.1  \\ 
\rowcolor{magenta!15}  Llama3.1 &   8B &   40.0 & 16.2 & 8.8 & 0.0 & 13.0 & 3.0 & 10.7 & 1.9 & 12.0 & 20.0 & 27.0 & 19.0 & 26.0 & 7.0 & 7.0 & 18.0 & 18.1 & 10.6        \\ 
\rowcolor{magenta!15} Deepseek-Coder & 6.7B &    37.1 & 21.9 & \underline{27.5} & \underline{7.8} & 20.0 & 3.0 & 4.9 & 1.9 & 16.0 & 17.0 & 30.0 & 20.0 & 30.0 & 8.0 & 12.0 & 19.0 & 22.2 & 12.3          \\ 
\rowcolor{magenta!15} Yi-Coder &   9B &  \underline{46.7} & 27.6 & 20.6 & 3.9 & 21.0 & 3.0 & \underline{12.6} & 2.9 & 17.0 & 20.0 & 38.0 & 28.0 & 38.0 & 8.0 & 17.0 & 23.0 & 26.4 & 14.6     \\ 
\rowcolor{magenta!15} Granite-Coder &   8B &  35.2 & 19.0 & 11.8 & 3.9 & 11.0 & 3.0 & 3.9 & 1.9 & 13.0 & 15.0 & 32.0 & 15.0 & 21.0 & 1.0 & 11.0 & 8.0 & 17.4 & 8.4      \\ 
\rowcolor{magenta!15} OpenCoder &     8B  &   38.1 & 24.8 & 21.6 & 5.9 & \underline{24.0} & \underline{4.0} & 5.8 & 1.9 & 16.0 & 17.0 & \underline{42.0} & 19.0 & 26.0 & 9.0 & 14.0 & 15.0 & 23.5 & 12.1          \\ 
\rowcolor{magenta!15} CodeQwen1.5 &   7B &   33.3 & 26.7 & 17.6 & 5.9 & 10.0 & \underline{4.0} & 4.9 & 1.9 & 14.0 & \underline{23.0} & 33.0 & 22.0 & 29.0 & 9.0 & 17.0 & 18.0 & 19.9 & 13.8            \\ 
\rowcolor{magenta!15} Qwen2.5-Coder &   7B & 41.9 & 27.6 & 25.5 & 3.9 & 16.0 & 3.0 & 11.7 & 2.9 & \underline{19.0} & 22.0 & 36.0 & \underline{30.0} & \underline{43.0} & 11.0 & 20.0 & \underline{28.0} & \underline{26.7} & \underline{16.0}      \\  
\rowcolor{magenta!15}  Qwen3 &   8B &  35.2 & \underline{33.3} & 13.7 & 7.8 & 12.0 & \underline{4.0} & 8.7 & 4.9 & 0.0 & 0.0 & 30.0 & 26.0 & 30.0 & \underline{19.0} & 9.0 & 27.0 & 17.4 & 15.3  \\ 
\rowcolor{magenta!15} Qwen3-think &   8B &        35.2 & 31.4 & 15.7 & 9.8 & 6.0 & \underline{4.0} & 1.9 & \underline{5.8} & 16.0 & 14.0 & 30.0 & 24.0 & 28.0 & 2.0 & \underline{22.0} & 14.0 & 19.4 & 13.2        \\ 
 \midrule
\multicolumn{20}{c}{\textit{14B+ Open-source LLMs}}      \\ \midrule
\rowcolor{violet!15} CodeLlama &   13B &  23.8 & 26.7 & 11.8 & 6.9 & 6.0 & \underline{5.0} & 1.0 & \underline{5.8} & 7.0 & 21.0 & 26.0 & 5.0 & 14.0 & 8.0 & 9.0 & 23.0 & 12.3 & 12.7         \\ 
\rowcolor{violet!15} Qwen2.5-Coder &   14B & 48.6 & 27.6 & \underline{27.5} & 7.8 & \underline{22.0} & 2.0 & \underline{18.4} & 2.9 & 20.0 & \underline{25.0} & 38.0 & 31.0 & 34.0 & 12.0 & 20.0 & 35.0 & 28.6 & 17.9 \\
\rowcolor{violet!15} Qwen3 &   14B &   36.2 & \underline{33.3} & 25.5 & 8.8 & 20.0 & 3.0 & 8.7 & 3.9 & 23.0 & 22.0 & 37.0 & 31.0 & 36.0 & \underline{20.0} & 18.0 & \underline{53.0} & 25.6 & \underline{21.9}        \\ 
\rowcolor{violet!15} Qwen3-think &   14B &   \underline{51.4} & 31.4 & 22.5 & \underline{15.7} & 16.0 & \underline{5.0} & 1.9 & 4.9 & \underline{30.0} & 24.0 & \underline{48.0} & \underline{33.0} & \underline{43.0} & 4.0 & \underline{24.0} & 17.0 & \underline{29.6} & 16.9       \\ 
\rowcolor{violet!15} Granite-Coder &   20B & 33.3 & 17.1 & 15.7 & 3.9 & 5.0 & \underline{5.0} & 4.9 & 1.9 & 15.0 & 15.0 & 26.0 & 19.0 & 22.0 & 4.0 & 11.0 & 9.0 & 16.7 & 9.4         \\ 
\midrule
\multicolumn{20}{c}{\textit{32B+ Open-source LLMs \& MoE LLMs}}      \\ \midrule
\rowcolor{red!15} Granite-Coder &   34B &  37.1 & 21.0 & 16.7 & 2.9 & 13.0 & 4.0 & 5.8 & 1.9 & 11.0 & 15.0 & 24.0 & 16.0 & 24.0 & 3.0 & 10.0 & 3.0 & 17.8 & 8.4        \\ 
\rowcolor{red!15} CodeLlama &   34B & 16.2 & 21.9 & 9.8 & 4.9 & 5.0 & 3.0 & 8.7 & 2.9 & 8.0 & 9.0 & 23.0 & 2.0 & 16.0 & 6.0 & 9.0 & 17.0 & 12.0 & 8.4        \\ 
\rowcolor{red!15} Deepseek-Coder & 33B &  41.9 & 22.9 & 33.3 & 4.9 & 22.0 & 4.0 & 7.8 & 1.9 & 24.0 & 20.0 & 33.0 & 18.0 & 37.0 & 7.0 & 19.0 & 20.0 & 27.3 & 12.3       \\ 
\rowcolor{red!15} Qwen2.5-Coder &    32B &   42.9 & 33.3 & 22.5 & 6.9 & 20.0 & 6.0 & 8.7 & 4.9 & 29.0 & 30.0 & 47.0 & 30.0 & 42.0 & 16.0 & 27.0 & 39.0 & 29.9 & 20.7           \\ 
\rowcolor{red!15} Llama3.1 &    70B &   42.9 & 33.3 & 22.5 & 6.9 & 20.0 & 6.0 & 8.7 & 4.9 & 29.0 & 30.0 & 47.0 & 30.0 & 42.0 & 16.0 & 27.0 & 39.0 & 29.9 & 20.7           \\ 
\rowcolor{red!15} Qwen3 & 32B &   49.5 & 33.3 & 25.5 & 8.8 & 17.0 & 7.0 & \underline{10.7} & 1.9 & 20.0 & 29.0 & 38.0 & \underline{34.0} & 41.0 & \underline{28.0} & 22.0 & \underline{52.0} & 28.0 & \underline{24.2}     \\ 
\rowcolor{red!15} Qwen3-think & 32B &   49.5 & 33.3 & 25.5 & 8.8 & 17.0 & 7.0 & \underline{10.7} & 1.9 & 20.0 & 29.0 & 38.0 & \underline{34.0} & 41.0 & \underline{28.0} & 22.0 & \underline{52.0} & 28.0 & \underline{24.2}     \\ 
\rowcolor{red!15} Qwen3 & 3/30B &  54.3 & 30.5 & 30.4 & 11.8 & 31.0 & 4.0 & 1.0 & 4.9 & 31.0 & 26.0 & 48.0 & 28.0 & 46.0 & 2.0 & 36.0 & 21.0 & 34.7 & 16.0           \\ 
\rowcolor{red!15} Qwen3-think & 3/30B & 35.2 & 33.3 & 16.7 & 11.8 & 14.0 & 6.0 & 1.9 & 4.9 & 32.0 & 30.0 & 45.0 & 33.0 & 46.0 & 4.0 & 34.0 & 22.0 & 28.0 & 18.1\\
\rowcolor{red!15} Qwen3 & 22B/235B &  50.5 & 33.3 & 28.4 & 9.8 & 14.0 & 2.0 & 1.9 & 4.9 & 20.0 & 24.0 & 43.0 & 29.0 & 42.0 & 16.0 & 26.0 & 35.0 & 28.3 & 19.3     \\ 
\rowcolor{red!15} Qwen3-think & 22B/235B &   48.6 & 32.4 & 21.6 & 10.8 & 20.0 & \underline{12.0} & 2.9 & \underline{6.8} & 21.0 & 15.0 & 37.0 & 22.0 & 47.0 & 4.0 & 35.0 & 15.0 & 29.1 & 14.8        \\ 
\rowcolor{red!15} Deepseek-V3 & 37/637B & 51.4 & 36.2 & 29.4 & 9.8 & 23.0 & 4.0 & 2.9 & 1.9 & 31.0 & 32.0 & 51.0 & 31.0 & 54.0 & 24.0 & 33.0 & 35.0 & 34.4 & 21.7     \\ 
\rowcolor{red!15} Deepseek-R1 & 37/637B &    58.1 & 37.1 & 37.3 & 12.7 & 38.0 & 5.0 & 4.9 & 5.8 & 46.0 & 35.0 & 55.0 & 28.0 & 61.0 & 6.0 & 45.0 & 20.0 & 43.1 & 18.8        \\ 
\rowcolor{red!15} \textbf{\baseline{}} &    32B &   58.3 & 38.2 & 39.3 & 13.5 & 30.0 & 12.0 & 10.7 & 6.8 & 44.0 & 38.0 & 47.0 & 38.0 & 48.0 & 30.0 & 45.0 & 52.0 & 40.3 & 34.2  \\ 
\bottomrule
\end{tabular}}
\vspace{-10pt}
\label{tab:ifevalcode_en}
\end{table*}

%% file: tables/table_zh.tex
\begin{table*}[]
\caption{Chinese evaluation results in \benchmark{} of 8 programming languages. `Corr..' denotes the correctness of the model-generated response, and `Instr.' denotes the instruction-following accuracy of the response. The underlined fonts denote the best performance in the same parameter range.}
\resizebox{1.0\textwidth}{!}{
\begin{tabular}{lc|cccccccccccccccccc}
\toprule
\multirow{2}{*}{Models}  & \multirow{2}{*}{Params} & \multicolumn{2}{c}{Python} & \multicolumn{2}{c}{Java} & \multicolumn{2}{c}{Cpp} & \multicolumn{2}{c}{C-sharp} & \multicolumn{2}{c}{Typescript} & \multicolumn{2}{c}{Javascript} & \multicolumn{2}{c}{Php} & \multicolumn{2}{c}{Shell} & \multicolumn{2}{c}{Avg.} \\ \cmidrule{3-20} 
 &    & Corr.       & Instr.       & Corr.      & Instr.      & Corr.      & Instr.     & Corr.        & Instr.       & Corr.         & Instr.         & Corr.         & Instr.         & Corr.      & Instr.     & Corr.       & Instr.      & Corr.      & Instr.      \\ \midrule
\multicolumn{20}{c}{\textit{Closed-source LLMs}}   \\ \midrule
\rowcolor{cyan!15} Claude-3.5-Sonnet  & \faLock{}   &  49.5 & 33.3 & \underline{38.2} & 11.8 & 23.0 & 7.0 & 14.6 & 2.9 & 20.0 & 29.0 & 45.0 & \underline{32.0} & 47.0 & 29.0 & 27.0 & 36.0 & 33.1 & 22.6\\  
\rowcolor{cyan!15} Claude-3.7-Sonnet  & \faLock{}   &  47.6 & 28.6 & 36.3 & 7.8 & 28.0 & 7.0 & 12.6 & 3.9 & 30.0 & \underline{30.0} & 51.0 & 30.0 & \underline{61.0} & 22.0 & 37.0 & 44.0 & \underline{37.9} & 21.6 \\  
\rowcolor{cyan!15} GPT-4o  & \faLock{}   &   42.9 & 39.0 & 24.5 & 7.8 & 22.0 & 7.0 & 14.6 & 3.9 & 27.0 & 26.0 & 46.0 & 23.0 & 51.0 & 27.0 & 30.0 & 42.0 & 32.2 & 22.0 \\
\rowcolor{cyan!15} GPT-4o-mini  & \faLock{}   &  45.7 & 28.6 & 21.6 & 9.8 & 19.0 & 3.0 & 12.6 & 2.9 & 20.0 & 25.0 & 40.0 & 28.0 & 43.0 & 25.0 & 20.0 & 31.0 & 27.8 & 19.1 \\
\rowcolor{cyan!15} GPT-4.1  & \faLock{}   &  51.4 & 39.0 & 35.3 & 11.8 & 25.0 & 7.0 & 0.0 & 2.9 & \underline{35.0} & 27.0 & 51.0 & 29.0 & 54.0 & 35.0 & 40.0 & 46.0 & 36.4 & \underline{24.7} \\
\rowcolor{cyan!15} GPT-4.1-mini  & \faLock{}   &  46.7 & 36.2 & \underline{38.2} & 6.9 & 18.0 & 3.0 & 0.0 & 2.9 & 32.0 & 28.0 & 50.0 & 28.0 & 46.0 & 33.0 & 43.0 & \underline{52.0} & 34.2 & 23.7 \\
\rowcolor{cyan!15} grok-3  & \faLock{}   &  30.5 & 24.8 & 24.5 & 9.8 & 2.0 & 5.0 & 0.0 & 2.9 & 4.0 & 16.0 & 46.0 & 23.0 & 37.0 & 32.0 & 16.0 & 35.0 & 20.0 & 18.5 \\
\rowcolor{cyan!15} grok-3-mini-fast  & \faLock{}   & 34.3 & 32.4 & 11.8 & 5.9 & 13.0 & \underline{13.0} & 1.9 & 2.9 & 26.0 & 21.0 & 44.0 & 27.0 & 33.0 & 4.0 & 12.0 & 20.0 & 22.0 & 15.8 \\
\midrule

\multicolumn{20}{c}{\textit{0.5B+ Open-source LLMs}}     \\ \midrule
\rowcolor{olive!15} Deepseek-Coder & 1.3B &   22.9 & 10.5 & 14.7 & 1.0 & 7.0 & 2.0 & 1.9 & 1.9 & 0.0 & 1.0 & 16.0 & 13.0 & 7.0 & 7.0 & 1.0 & 2.0 & 8.9 & 4.8         \\ 
\rowcolor{olive!15} Qwen2.5-Coder &   0.5B &  15.2 & 16.2 & 10.8 & 2.0 & 8.0 & 3.0 & 7.8 & 1.9 & 1.0 & 14.0 & 13.0 & 15.0 & 7.0 & 7.0 & 1.0 & 11.0 & 8.0 & 8.8  \\ 
\rowcolor{olive!15} Qwen2.5-Coder &   1.5B &  28.6 & 21.9 & 9.8 & 2.9 & 2.0 & 5.0 & 0.0 & 1.9 & 6.0 & 20.0 & 28.0 & \underline{23.0} & 18.0 & 11.0 & 4.0 & 24.0 & 12.1 & 13.7 \\ 
\rowcolor{olive!15} Qwen2.5-Coder &   3B  & 31.4 & 20.0 & 13.7 & 3.9 & 12.0 & 5.0 & 11.7 & 2.9 & 7.0 & 19.0 & \underline{31.0} & \underline{23.0} & 27.0 & 12.0 & 10.0 & 30.0 & \underline{18.0} & 14.4      \\ 
\rowcolor{olive!15} Granite-Coder &   3B &  25.7 & 21.0 & 10.8 & 4.9 & 7.0 & 3.0 & 12.6 & \underline{3.9} & 6.0 & 14.0 & 18.0 & 11.0 & 15.0 & 1.0 & 5.0 & 12.0 & 12.6 & 8.9 \\ 
\rowcolor{olive!15} OpenCoder & 1.5B & \underline{39.0} & 16.2 & \underline{21.6} & 3.9 & \underline{20.0} & 2.0 & 6.8 & 1.9 & 10.0 & 14.0 & 18.0 & 14.0 & 21.0 & 8.0 & 7.0 & 17.0 & \underline{18.0} & 9.6   \\ 
\rowcolor{olive!15} Yi-Coder &   1.5B &  34.3 & 17.1 & 15.7 & 1.0 & 16.0 & 5.0 & \underline{14.6} & 1.9 & 10.0 & 11.0 & 22.0 & 21.0 & 15.0 & 7.0 & 2.0 & 13.0 & 16.3 & 9.6   \\ 
\rowcolor{olive!15} Qwen3 & 0.6B & 11.4 & 21.0 & 2.9 & 3.9 & 1.0 & \underline{6.0} & 0.0 & 1.9 & 0.0 & 25.0 & 13.0 & 19.0 & 8.0 & 15.0 & 0.0 & 22.0 & 4.6 & 14.2    \\ 
\rowcolor{olive!15} Qwen3-think & 0.6B & 10.5 & 10.5 & 2.9 & 3.9 & 0.0 & 2.0 & 0.0 & 1.9 & 1.0 & 10.0 & 9.0 & 8.0 & 6.0 & 1.0 & 0.0 & 7.0 & 3.7 & 5.6  \\ 
\rowcolor{olive!15} Qwen3 &   1.7B  & 27.6 & \underline{29.5} & 6.9 & 3.9 & 4.0 & 5.0 & 1.0 & 1.9 & 9.0 & 25.0 & 19.0 & 19.0 & 15.0 & 13.0 & 5.0 & 31.0 & 11.0 & 16.0 \\ 
\rowcolor{olive!15} Qwen3-think &   1.7B  & 21.0 & 14.3 & 2.9 & 2.9 & 0.0 & 0.0 & 0.0 & 1.0 & 10.0 & 1.0 & 14.0 & 3.0 & 21.0 & 2.0 & 3.0 & 3.0 & 9.0 & 3.5  \\        
\rowcolor{olive!15} Qwen3 &   4B  & 24.8 & 30.5 & 6.9 & \underline{7.8} & 5.0 & 4.0 & 2.9 & \underline{3.9} & \underline{14.0} & \underline{26.0} & 28.0 & 31.0 & \underline{29.0} & \underline{20.0} & 8.0 & \underline{47.0} & 14.8 & \underline{21.2} \\ 
\rowcolor{olive!15} Qwen3-think &   4B  & 24.8 & 7.6 & 4.9 & 1.0 & 1.0 & 1.0 & 0.0 & 0.0 & 9.0 & 5.0 & 16.0 & 3.0 & 21.0 & 1.0 & \underline{14.0} & 6.0 & 11.4 & 3.1  \\ \midrule

\multicolumn{20}{c}{\textit{6B+ Open-source LLMs}}       \\ \midrule
\rowcolor{magenta!15}  CodeLlama &   7B &    0.0 & 1.9 & 5.9 & 1.0 & 1.0 & 3.0 & 0.0 & 2.9 & 5.0 & 2.0 & 23.0 & 0.0 & 1.0 & 2.0 & 2.0 & 17.0 & 4.7 & 3.7  \\ 
\rowcolor{magenta!15}  Llama3.1 &   8B &  35.2 & 21.9 & 9.8 & 2.0 & 13.0 & 2.0 & \underline{15.5} & 1.9 & 10.0 & 19.0 & 24.0 & 23.0 & 24.0 & 14.0 & 5.0 & 24.0 & 17.2 & 13.5      \\ 
\rowcolor{magenta!15} Deepseek-Coder & 6.7B &   38.1 & 23.8 & 20.6 & 3.9 & 16.0 & 3.0 & 2.9 & 1.9 & \underline{18.0} & 16.0 & 34.0 & 17.0 & 25.0 & 10.0 & 16.0 & 23.0 & 21.4 & 12.3        \\ 
\rowcolor{magenta!15} Yi-Coder &   9B & \underline{43.8} & 24.8 & \underline{31.4} & \underline{5.9} & \underline{21.0} & 2.0 & 7.8 & 1.9 & \underline{18.0} & 16.0 & 33.0 & 24.0 & \underline{46.0} & 11.0 & 14.0 & 23.0 & \underline{26.9} & 13.6  \\ 
\rowcolor{magenta!15} Granite-Coder &   8B & 32.4 & 19.0 & 11.8 & 3.9 & 14.0 & 4.0 & 6.8 & 1.9 & 9.0 & 13.0 & 29.0 & 19.0 & 9.0 & 3.0 & 10.0 & 16.0 & 15.3 & 10.0   \\ 
\rowcolor{magenta!15} OpenCoder &     8B  &   40.0 & 25.7 & 17.6 & 4.9 & 16.0 & \underline{5.0} & 4.9 & 1.9 & 15.0 & 19.0 & 30.0 & 19.0 & 32.0 & 12.0 & 13.0 & 26.0 & 21.1 & 14.2         \\ 
\rowcolor{magenta!15} CodeQwen1.5 &   7B &  41.0 & 21.9 & 15.7 & 4.9 & 12.0 & 4.0 & 9.7 & 1.9 & 12.0 & \underline{23.0} & 33.0 & 25.0 & 27.0 & 11.0 & 13.0 & 23.0 & 20.5 & 14.3        \\ 
\rowcolor{magenta!15} Qwen2.5-Coder &   7B & 42.9 & 27.6 & 21.6 & 4.9 & 11.0 & \underline{5.0} & 6.8 & 2.9 & 16.0 & 21.0 & \underline{39.0} & 29.0 & 36.0 & 11.0 & \underline{21.0} & \underline{32.0} & 24.3 & \underline{16.7}     \\  
\rowcolor{magenta!15}  Qwen3 &   8B &  36.2 & \underline{32.4} & 11.8 & \underline{5.9} & 7.0 & \underline{5.0} & 5.8 & \underline{3.9} & 0.0 & 0.0 & 29.0 & 32.0 & \underline{32.0} & \underline{27.0} & 6.0 & 18.0 & 16.0 & 15.6 \\ 
\rowcolor{magenta!15} Qwen3-think &   8B &  28.6 & 12.4 & 8.8 & 2.9 & 3.0 & 1.0 & 1.9 & 1.0 & 14.0 & 2.0 & 26.0 & 6.0 & 27.0 & 0.0 & 15.0 & 6.0 & 15.6 & 4.0       \\ 
 \midrule

\multicolumn{20}{c}{\textit{14B+ Open-source LLMs}}      \\ \midrule
\rowcolor{violet!15} CodeLlama &   13B &  19.0 & 22.9 & 10.8 & 3.9 & 7.0 & \underline{4.0} & 4.9 & 2.9 & 7.0 & 24.0 & 24.0 & 8.0 & 22.0 & 7.0 & 4.0 & 22.0 & 12.3 & 11.9        \\ 
\rowcolor{violet!15} Qwen2.5-Coder &   14B &47.6 & 29.5 & \underline{26.5} & \underline{8.8} & \underline{22.0} & 3.0 & \underline{19.4} & 2.9 & \underline{24.0} & \underline{32.0} & \underline{41.0} & 30.0 & 39.0 & 15.0 & 19.0 & 36.0 & \underline{29.9} & 19.6\\
\rowcolor{violet!15} Qwen3 &   14B &  39.0 & \underline{32.4} & 19.6 & \underline{8.8} & 17.0 & 3.0 & 10.7 & \underline{4.9} & 23.0 & 31.0 & \underline{41.0} & \underline{33.0} & \underline{40.0} & \underline{23.0} & 18.0 & \underline{53.0} & 26.0 & \underline{23.6}    \\ 
\rowcolor{violet!15} Qwen3-think &   14B &  \underline{49.5} & 26.7 & 16.7 & 2.9 & 10.0 & 1.0 & 2.9 & 1.0 & 23.0 & 8.0 & 34.0 & 6.0 & 37.0 & 1.0 & \underline{22.0} & 10.0 & 24.4 & 7.2    \\ 
\rowcolor{violet!15} Granite-Coder &   20B & 30.5 & 21.9 & 15.7 & 2.9 & 4.0 & 4.0 & 6.8 & 3.9 & 13.0 & 13.0 & 25.0 & 19.0 & 26.0 & 3.0 & 4.0 & 10.0 & 15.7 & 9.8      \\ 
\midrule
\multicolumn{20}{c}{\textit{32B+ Open-source LLMs \& MoE LLMs}}      \\ \midrule
\rowcolor{red!15} Granite-Coder &   34B & 34.3 & 23.8 & 12.7 & 4.9 & 10.0 & 4.0 & 5.8 & 1.9 & 16.0 & 15.0 & 29.0 & 23.0 & 24.0 & 4.0 & 5.0 & 9.0 & 17.2 & 10.7   \\ 
\rowcolor{red!15} CodeLlama &  34B & 4.8 & 15.2 & 5.9 & 2.0 & 15.0 & 3.0 & 12.6 & 1.9 & 10.0 & 6.0 & 18.0 & 0.0 & 17.0 & 10.0 & 3.0 & 22.0 & 10.7 & 7.5       \\ 
\rowcolor{red!15} Deepseek-Coder & 33B &  44.8 & 21.0 & 26.5 & 4.9 & 24.0 & \underline{7.0} & 7.8 & 1.9 & 21.0 & 18.0 & 35.0 & 20.0 & 30.0 & 14.0 & 20.0 & 26.0 & 26.2 & 14.1    \\ 
\rowcolor{red!15} Qwen2.5-Coder &    32B &  43.8 & 34.3 & 28.4 & 8.8 & 19.0 & 2.0 & \underline{12.6} & 5.8 & 25.0 & 29.0 & 41.0 & 34.0 & 44.0 & 24.0 & 23.0 & 38.0 & 29.6 & 22.0      \\ 
\rowcolor{red!15} Llama3.1 &    70B &   47.6 & 22.9 & 14.7 & 2.0 & 19.0 & 3.0 & 11.7 & 2.9 & 19.0 & 24.0 & 35.0 & 28.0 & 39.0 & 19.0 & 21.0 & 40.0 & 25.9 & 17.7      \\ 
\rowcolor{red!15} Qwen3 & 32B &  42.9 & \underline{41.0} & 24.5 & 7.8 & 17.0 & \underline{7.0} & 9.7 & 3.9 & 27.0 & 25.0 & 38.0 & 31.0 & 40.0 & 23.0 & 19.0 & 51.0 & 27.3 & 23.7   \\ 
\rowcolor{red!15} Qwen3-think & 32B &  54.3 & 19.0 & 30.4 & 5.9 & 24.0 & 2.0 & 7.8 & 1.9 & 26.0 & 2.0 & 42.0 & 7.0 & 50.0 & 1.0 & 30.0 & 6.0 & 33.1 & 5.7  \\ 
\rowcolor{red!15} Qwen3 & 3/30B & 43.8 & 33.3 & 28.4 & 11.8 & 19.0 & 6.0 & 5.8 & 5.8 & 24.0 & 31.0 & 42.0 & 31.0 & 38.0 & 29.0 & 21.0 & 48.0 & 27.8 & 24.4    \\ 
\rowcolor{red!15} Qwen3-think & 3/30B & 42.9 & 21.0 & 16.7 & 2.9 & 12.0 & 0.0 & 2.9 & 2.9 & 21.0 & 4.0 & 41.0 & 10.0 & 40.0 & 1.0 & 22.0 & 7.0 & 24.8 & 6.2 \\
\rowcolor{red!15} Qwen3 & 22B/235B & 46.7 & 38.1 & 22.5 & 8.8 & 12.0 & 3.0 & 1.0 & 2.9 & 25.0 & 24.0 & 44.0 & 30.0 & 47.0 & 22.0 & 20.0 & 38.0 & 27.3 & 20.9   \\ 
\rowcolor{red!15} Qwen3-think & 22B/235B &   46.7 & 11.4 & 26.5 & 2.0 & 26.0 & 1.0 & 1.0 & 0.0 & 32.0 & 1.0 & 43.0 & 2.0 & 43.0 & 1.0 & 36.0 & 2.0 & 31.7 & 2.6    \\ 
\rowcolor{red!15} Deepseek-V3 & 37/671B & 50.5 & 35.2 & 32.4 & 9.8 & 21.0 & 5.0 & 4.9 & 1.9 & 32.0 & 34.0 & 51.0 & 25.0 & 51.0 & 29.0 & 34.0 & 38.0 & 34.6 & 22.2     \\ 
\rowcolor{red!15} Deepseek-R1 & 37/671B &    \underline{58.1} & 37.1 & 37.3 & 12.7 & \underline{38.0} & 5.0 & 4.9 & 5.8 & 46.0 & \underline{35.0} & \underline{55.0} & 28.0 & \underline{61.0} & 6.0 & 45.0 & 20.0 & \underline{43.1} & 18.8        \\ 
\rowcolor{red!15} \textbf{\baseline{}} &    32B &   54.8 & 39.2 & \underline{40.3} & \underline{13.8} & 34.0 & \underline{7.0} & 11.7 & \underline{6.8} & \underline{48.0} & \underline{35.0} & 49.0 & \underline{40.0} & 48.0 & \underline{32.0} & \underline{51.0} & \underline{55.0} & 42.1 & \underline{28.6}  \\ 
\bottomrule
\end{tabular}}
\label{tab:ifevalcode_zh}
\end{table*}

%% file: appendix.tex
\appendix
\section{List of Large Language Models}
%%%%%%%%%%%%%%%

Table~\ref{tab:model_list} lists all the large language models evaluated in this study and their download links; we then systematically evaluate the performance of these models on the \benchmark{} benchmark.

\begin{table*}[h]
\centering
\caption{Evaluation Model List}
\resizebox{1.0\textwidth}{!}{
\begin{tabular}{ll}
\toprule
Model & Download Link \\ 
\midrule
Claude-3.7-Sonnet & \url{https://www.anthropic.com/news/claude-3-7-sonnet}     \\ 
Claude-3.5-Sonnet & \url{https://www.anthropic.com/news/claude-3-5-sonnet}     \\ 
GPT-4o & \url{https://platform.openai.com/docs/models/gpt-4o} \\
GPT-4o-mini & \url{https://platform.openai.com/docs/models/gpt-4o-mini} \\
GPT-4.1 & \url{https://platform.openai.com/docs/models/gpt-4.1} \\
GPT-4.1-mini & \url{https://platform.openai.com/docs/models/gpt-4.1-mini} \\
o1-mini & \url{https://platform.openai.com/docs/models/o1-mini} \\
o3-mini & \url{https://platform.openai.com/docs/models/o3-mini} \\
o4-mini & \url{https://platform.openai.com/docs/models/o4-mini} \\
grok-3 & \url{https://docs.x.ai/docs/models\#models-and-pricing} \\
grok-3-mini-fast & \url{https://docs.x.ai/docs/models\#models-and-pricing} \\
Deepseek-Coder-1.3B & \url{https://huggingface.co/deepseek-ai/deepseek-coder-1.3b-instruct} \\
Qwen2.5-Coder-0.5B & \url{https://huggingface.co/Qwen/Qwen2.5-Coder-0.5B} \\
Qwen2.5-Coder-1.5B & \url{https://huggingface.co/Qwen/Qwen2.5-Coder-1.5B} \\
Qwen2.5-Coder-3B & \url{https://huggingface.co/Qwen/Qwen2.5-Coder-3B} \\
Granite-Coder-3B & \url{https://huggingface.co/ibm-granite/granite-3b-code-base-128k} \\
OpenCoder-1.5B & \url{https://huggingface.co/infly/OpenCoder-1.5B-Instruct} \\
Qwen3-0.6B      & \url{https://huggingface.co/Qwen/Qwen3-0.6B}     \\
Qwen3-1.7B      & \url{https://huggingface.co/Qwen/Qwen3-1.7B}     \\
Qwen3-4B & \url{https://huggingface.co/Qwen/Qwen3-4B}    \\ 
CodeLlama-7B & \url{https://huggingface.co/meta-llama/CodeLlama-7b-Instruct-hf} \\ 
Deepseek-Coder-6.7B & \url{https://huggingface.co/deepseek-ai/deepseek-coder-6.7b-instruct} \\
Yi-Coder-9B & \url{https://huggingface.co/01-ai/Yi-Coder-9B-Chat} \\
Granite-Coder-8B & \url{https://huggingface.co/ibm-granite/granite-8b-code-instruct-128k} \\
OpenCoder-8B & \url{https://huggingface.co/infly/OpenCoder-8B-Instruct} \\
CoderQwen1.5-7B & \url{https://huggingface.co/Qwen/CodeQwen1.5-7B-Chat} \\
Qwen2.5-Coder-7B & \url{https://huggingface.co/Qwen/Qwen2.5-Coder-7B} \\
Qwen3-8B        & \url{https://huggingface.co/Qwen/Qwen3-8B}     \\
Qwen2.5-Coder-14B & \url{https://huggingface.co/Qwen/Qwen2.5-Coder-14B} \\
Qwen3-14B        & \url{https://huggingface.co/Qwen/Qwen3-14B}     \\
Granite-Coder-20B & \url{https://huggingface.co/ibm-granite/granite-20b-code-instruct-8k} \\
Granite-Coder-34B & \url{https://huggingface.co/ibm-granite/granite-34b-code-instruct-8k} \\
Deepseek-Coder-33B & \url{https://huggingface.co/deepseek-ai/deepseek-coder-33b-instruct} \\
Qwen2.5-Coder-32B & \url{https://huggingface.co/Qwen/Qwen2.5-Coder-32B} \\
Qwen3-32B       & \url{https://huggingface.co/Qwen/Qwen3-32B}     \\
Qwen3-30B-A3B & \url{https://huggingface.co/Qwen/Qwen3-30B-A3B} \\
Qwen3-235B-A22B & \url{https://huggingface.co/Qwen/Qwen3-235B-A22B} \\
DeepSeek-V3 & \url{https://huggingface.co/deepseek-ai/DeepSeek-V3} \\
DeepSeek-R1 & \url{https://huggingface.co/deepseek-ai/DeepSeek-R1} \\
\bottomrule
\end{tabular}}
\label{tab:model_list}
\end{table*}
%%%%%%%%%%%%%%%

\section{Limitations}
\paragraph{Programming Languages} To ensure a more comprehensive evaluation of multilingual controlled code generation, it would be beneficial to expand \benchmark{} to include additional programming languages, particularly those that are less commonly used or serve niche applications, as the current coverage of eight languages may not fully represent the diversity and specialized use cases present in real-world software development.

\paragraph{Complex Controlled Code Generation} While \benchmark{} aims to simulate realistic controlled code generation scenarios, \benchmark{} ignores the complex controlled code generation and lacks repository-based evaluation.

\section{Ethical Considerations}
\subsection{Potential Risks}
\benchmark{} serves as a comprehensive assessment tool for controlled code generation. However, improper use may lead to risks such as incorrect program analysis, faulty repairs, or even system failures. To mitigate these risks, we strongly recommend conducting evaluations within a sandbox environment, which effectively isolates potential hazards and safeguards both the accuracy of the assessment and the stability of the underlying system.

\section{Human Annotation}
\paragraph{Annotators} To construct \benchmark{}, we developed a rigorous and systematic human annotation process, guided by meticulously designed annotation protocols, to ensure the accuracy, consistency, and high quality of questions. This process integrates multiple quality control measures at every stage to maintain reliability and precision across all collected data. We recruited $6$ programmers with computer science degrees equipped with strong computer science fundamentals.

\section{Analysis of Instruction Constraints}
Illustrated in Figure~\ref{fig:instruction_types}, data annotation was based on diverse, real-world constraint types frequently encountered in software engineering. In our annotation, key challenge clusters for LLMs are:

\begin{itemize}
\item \textbf{Naming constraints} (e.g., ``use camelCase'' or ``no abbreviation''): Models often ignore subtle naming conventions unless strongly prompted.
\item \textbf{Syntax restrictions} (e.g., ``no recursion'', ``use list comprehension''): Many LLMs hallucinate or use default constructs, especially in multi-turn or long/intricate tasks.
\item \textbf{Variable constraints} (e.g., variable reuse/uniqueness): LLMs may create unnecessary variables or reuse names incorrectly.
\item \textbf{Style/format constraints} (e.g., number of lines, comment placement): These are often missed as they are not strongly encoded in pretraining.
Performance and resource constraints (e.g., avoid O($n^2$) complexity): LLMs do not reliably estimate algorithmic efficiency.
\end{itemize}

The most challenging for LLMs are ``style constraints,'' ``naming constraints,'' and ``performance/resource constraints,'' as these are not always explicit in training corpora and often conflict with models’ default generations. Performance varies by LLM family; for example, Claude Series models are slightly better, but none are consistently reliable across all constraint categories.

\section{Related Work}
\paragraph{Code Large Language Models.}
Large language models (LLMs) designed for coding tasks have demonstrated exceptional capabilities in code generation, debugging, translation, and other essential functions for modern software engineering~\citep{Chen2021Evaluating,claude,gpt4,unicoder}, with numerous benchmarks developed to evaluate these capabilities, though many focus on a limited selection of programming languages like Python and Java~\citep{codegeex,mbpp,livecodebench}. Recent advancements in code LLMs, such as Code Llama~\citep{codellama}, DeepSeek-Coder~\citep{deepseek_coder}, OpenCoder~\citep{opencoder}, and Qwen2.5-Coder~\citep{qwen25coder}, have significantly improved multilingual code generation and debugging, evaluated using benchmarks like MultiPL-E~\citep{multiple}, McEval~\citep{mceval}, and MdEval~\citep{mdevl}. Code generation, a fundamental task for code LLMs, requires interpreting natural language descriptions and generating corresponding code snippets~\citep{cruxeval,ds1000,evalplus,autokaggle}, with benchmarks proposed for code translation~\citep{codetransocean}, code completion~\citep{fim,repocoder}, and structured data understanding~\citep{tablebench,tablegpt2}. 

\paragraph{Instruction Following.}
Automatic program debugging holds substantial practical value. With the emergence of LLM capabilities, a growing number of individuals are utilizing LLMs for code debugging, leading to extensive research in this field.
Code Debugging includes several tasks such as bug or vulnerability detection~\citep{pradel2018deepbugs,allamanis2021self,yuan2023evaluating,zhang2024prompt,zhong2024advancing}, fuzz test~\citep{deng2023large,xia2024fuzz4all,yang2024fuzzcoder}, program repair~\citep{wen2024fixing,quixbugs,EvalGPTFix,runbugrun,gu2024counterfeit,tambon2024bugs,wang2024large}, GitHub issues auto resolving~\citep{swe_bench,coder,magis}.
To effectively assess the code debugging capabilities of LLMs, several benchmark tests have been introduced~\citep{prenner2022can,sobania2023analysis,xia2023conversational,EvalGPTFix,debugbench,debugeval}. Notably, DebugBench~\citep{debugbench} provides a comprehensive classification of error types and analyzes the debugging capabilities of LLMs based on these categories. Similarly, DebugEval~\citep{debugeval} has designed various debugging-related tasks to evaluate LLM performance across different task dimensions. However, these studies focus on 1 to 3 languages. In reality, there are significant differences in code errors between languages, leading to numerous language-specific errors. To address this gap, we propose \benchmark{}, a comprehensive code debugging benchmark covering 20 languages, aiming to assess LLM debugging capabilities from a broader perspective.

%% file: neurips.bib
@article{qwencoder,
  title={Qwen2. 5-coder technical report},
  author={Hui, Binyuan and Yang, Jian and Cui, Zeyu and Yang, Jiaxi and Liu, Dayiheng and Zhang, Lei and Liu, Tianyu and Zhang, Jiajun and Yu, Bowen and Dang, Kai and others},
  journal={arXiv preprint arXiv:2409.12186},
  year={2024}
}

@article{qwen25coder,
  title={Qwen2. 5-Coder Technical Report},
  author={Hui, Binyuan and Yang, Jian and Cui, Zeyu and Yang, Jiaxi and Liu, Dayiheng and Zhang, Lei and Liu, Tianyu and Zhang, Jiajun and Yu, Bowen and Dang, Kai and others},
  journal={arXiv preprint arXiv:2409.12186},
  year={2024}
}

@article{mdevl,
  title={MdEval: Massively Multilingual Code Debugging},
  author={Liu, Shukai and Chai, Linzheng and Yang, Jian and Shi, Jiajun and Zhu, He and Wang, Liran and Jin, Ke and Zhang, Wei and Zhu, Hualei and Guo, Shuyue and others},
  journal={arXiv preprint arXiv:2411.02310},
  year={2024}
}

@article{autokaggle,
  title={AutoKaggle: A Multi-Agent Framework for Autonomous Data Science Competitions},
  author={Li, Ziming and Zang, Qianbo and Ma, David and Guo, Jiawei and Zheng, Tianyu and Niu, Xinyao and Yue, Xiang and Wang, Yue and Yang, Jian and Liu, Jiaheng and others},
  journal={arXiv preprint arXiv:2410.20424},
  year={2024}
}

@article{tablegpt2,
  title={TableGPT2: A Large Multimodal Model with Tabular Data Integration},
  author={Su, Aofeng and Wang, Aowen and Ye, Chao and Zhou, Chen and Zhang, Ga and Zhu, Guangcheng and Wang, Haobo and Xu, Haokai and Chen, Hao and Li, Haoze and others},
  journal={arXiv preprint arXiv:2411.02059},
  year={2024}
}

@article{tablebench,
  title={TableBench: A Comprehensive and Complex Benchmark for Table Question Answering},
  author={Wu, Xianjie and Yang, Jian and Chai, Linzheng and Zhang, Ge and Liu, Jiaheng and Du, Xinrun and Liang, Di and Shu, Daixin and Cheng, Xianfu and Sun, Tianzhen and others},
  journal={arXiv preprint arXiv:2408.09174},
  year={2024}
}

@Article{cruxeval,
  author       = {Gu, Alex and Rozi{\`e}re, Baptiste and Leather, Hugh and Solar-Lezama, Armando and Synnaeve, Gabriel and Wang, Sida I},
  year         = {2024},
  journaltitle = {arXiv preprint arXiv:2401.03065},
  title        = {Cruxeval: A benchmark for code reasoning, understanding and execution},
}

@ARTICLE{multiple,
  author={Cassano, Federico and Gouwar, John and Nguyen, Daniel and Nguyen, Sydney and Phipps-Costin, Luna and Pinckney, Donald and Yee, Ming-Ho and Zi, Yangtian and Anderson, Carolyn Jane and Feldman, Molly Q and Guha, Arjun and Greenberg, Michael and Jangda, Abhinav},
  journal={IEEE Transactions on Software Engineering}, 
  title={MultiPL-E: A Scalable and Polyglot Approach to Benchmarking Neural Code Generation}, 
  year={2023},
  volume={49},
  number={7},
  pages={3675-3691},
  doi={10.1109/TSE.2023.3267446}
}

@Article{codellama,
  author       = {Roziere, Baptiste and Gehring, Jonas and Gloeckle, Fabian and Sootla, Sten and Gat, Itai and Tan, Xiaoqing Ellen and Adi, Yossi and Liu, Jingyu and Remez, Tal and Rapin, J{\'e}r{\'e}my and others},
  year         = {2023},
  journaltitle = {arXiv preprint arXiv:2308.12950},
  title        = {Code llama: Open foundation models for code},
}

@article{chen2021evaluating,
 author = {Chen, Mark and Tworek, Jerry and Jun, Heewoo and Yuan, Qiming and Pinto, Henrique Ponde de Oliveira and Kaplan, Jared and Edwards, Harri and Burda, Yuri and Joseph, Nicholas and Brockman, Greg and others},
 journal = {ArXiv preprint},
 title = {Evaluating large language models trained on code},
 url = {https://arxiv.org/abs/2107.03374},
 volume = {abs/2107.03374},
 year = {2021}
}

@article{fim,
  title={Efficient Training of Language Models to Fill in the Middle},
  author={Bavarian, Mohammad and Jun, Heewoo and Tezak, Nikolas and Schulman, John and McLeavey, Christine and Tworek, Jerry and Chen, Mark},
  journal={arXiv preprint arXiv:2207.14255},
  year={2022},
  url={https://arxiv.org/abs/2207.14255},
}

@inproceedings{code_bert,
    title = "CodeBERT: A Pre-Trained Model for Programming and Natural Languages",
    author = "Feng, Zhangyin  and
      Guo, Daya  and
      Tang, Duyu  and
      Duan, Nan  and
      Feng, Xiaocheng  and
      Gong, Ming  and
      Shou, Linjun  and
      Qin, Bing  and
      Liu, Ting  and
      Jiang, Daxin  and
      Zhou, Ming",
    editor = "Cohn, Trevor  and
      He, Yulan  and
      Liu, Yang",
    booktitle = "Findings of the Association for Computational Linguistics: EMNLP 2020",
    month = nov,
    year = "2020",
    address = "Online",
    publisher = "Association for Computational Linguistics",
    url = "https://aclanthology.org/2020.findings-emnlp.139",
    doi = "10.18653/v1/2020.findings-emnlp.139",
    pages = "1536--1547",
}

@article{deepseek_coder,
  title={DeepSeek-Coder: When the Large Language Model Meets Programming -- The Rise of Code Intelligence},
  author={Guo, Daya and Zhu, Qihao and Yang, Dejian and Xie, Zhenda and Dong, Kai and Zhang, Wentao and Chen, Guanting and Bi, Xiao and Wu, Y and Li, YK and others},
  journal={arXiv preprint arXiv:2401.14196},
  year={2024},
  url={https://arxiv.org/abs/2401.14196},
}

@article{codegeex,
  author       = {Qinkai Zheng and
                  Xiao Xia and
                  Xu Zou and
                  Yuxiao Dong and
                  Shan Wang and
                  Yufei Xue and
                  Zihan Wang and
                  Lei Shen and
                  Andi Wang and
                  Yang Li and
                  Teng Su and
                  Zhilin Yang and
                  Jie Tang},
  title        = {CodeGeeX: {A} Pre-Trained Model for Code Generation with Multilingual
                  Evaluations on HumanEval-X},
  journal      = {arXiv preprint arXiv:2303.17568},
  volume       = {abs/2303.17568},
  year         = {2023},
  url          = {https://doi.org/10.48550/arXiv.2303.17568},
  doi          = {10.48550/ARXIV.2303.17568},
  eprinttype    = {arXiv},
  eprint       = {2303.17568},
}

@article{Qwen,
  author       = {Jinze Bai and
                  Shuai Bai and
                  Yunfei Chu and
                  Zeyu Cui and
                  Kai Dang and
                  Xiaodong Deng and
                  Yang Fan and
                  Wenbin Ge and
                  Yu Han and
                  Fei Huang and
                  Binyuan Hui and
                  Luo Ji and
                  Mei Li and
                  Junyang Lin and
                  Runji Lin and
                  Dayiheng Liu and
                  Gao Liu and
                  Chengqiang Lu and
                  Keming Lu and
                  Jianxin Ma and
                  Rui Men and
                  Xingzhang Ren and
                  Xuancheng Ren and
                  Chuanqi Tan and
                  Sinan Tan and
                  Jianhong Tu and
                  Peng Wang and
                  Shijie Wang and
                  Wei Wang and
                  Shengguang Wu and
                  Benfeng Xu and
                  Jin Xu and
                  An Yang and
                  Hao Yang and
                  Jian Yang and
                  Shusheng Yang and
                  Yang Yao and
                  Bowen Yu and
                  Hongyi Yuan and
                  Zheng Yuan and
                  Jianwei Zhang and
                  Xingxuan Zhang and
                  Yichang Zhang and
                  Zhenru Zhang and
                  Chang Zhou and
                  Jingren Zhou and
                  Xiaohuan Zhou and
                  Tianhang Zhu},
  title        = {Qwen Technical Report},
  journal      = {arXiv preprint arXiv:2309.16609},
  volume       = {abs/2309.16609},
  year         = {2023},
  url          = {https://arxiv.org/abs/2309.16609}
}

@inproceedings{CodeTransOcean,
    title = "{C}ode{T}rans{O}cean: A Comprehensive Multilingual Benchmark for Code Translation",
    author = "Yan, Weixiang  and
      Tian, Yuchen  and
      Li, Yunzhe  and
      Chen, Qian  and
      Wang, Wen",
    editor = "Bouamor, Houda  and
      Pino, Juan  and
      Bali, Kalika",
    booktitle = "Findings of the Association for Computational Linguistics: EMNLP 2023",
    month = dec,
    year = "2023",
    address = "Singapore",
    publisher = "Association for Computational Linguistics",
    url = "https://aclanthology.org/2023.findings-emnlp.337",
    doi = "10.18653/v1/2023.findings-emnlp.337",
    pages = "5067--5089",
}

@article{evalplus,
  author       = {Jiawei Liu and
                  Chunqiu Steven Xia and
                  Yuyao Wang and
                  Lingming Zhang},
  title        = {Is Your Code Generated by ChatGPT Really Correct? Rigorous Evaluation of Large Language Models for Code Generation},
  volume       = {abs/2305.01210},
  year         = {2023},
  journal={arXiv preprint arXiv:2305.01210},
  url={https://arxiv.org/abs/2305.01210},
}

@article{code_llama,
  title={Code Llama: Open Foundation Models for Code},
  author={Rozi{\`e}re, Baptiste and Gehring, Jonas and Gloeckle, Fabian and Sootla, Sten and Gat, Itai and Tan, Xiaoqing Ellen and Adi, Yossi and Liu, Jingyu and Remez, Tal and Rapin, J{\'e}r{\'e}my and others},
  journal={arXiv preprint arXiv:2308.12950},
  year={2023},
  url={https://arxiv.org/abs/2308.12950},
}

@misc{claude,
  title = {Introducing {Claude}},
  author = {Anthropic},
  institution = {Anthropic},
  url = {https://www.anthropic.com/index/introducing-claude},
  year={2023}
}

@article{llama2,
  title={Llama 2: Open foundation and fine-tuned chat models},
  author={Touvron, Hugo and Martin, Louis and Stone, Kevin and Albert, Peter and Almahairi, Amjad and Babaei, Yasmine and Bashlykov, Nikolay and Batra, Soumya and Bhargava, Prajjwal and Bhosale, Shruti and others},
  journal={arXiv preprint arXiv:2307.09288},
  year={2023}
}

@article{gpt4,
  title={GPT-4 Technical Report},
  author={OpenAI},
  journal={arXiv preprint arXiv:2303.08774},
  year={2023},
  url={https://arxiv.org/abs/2303.08774},
}

@article{opencoder,
  title={Opencoder: The open cookbook for top-tier code large language models},
  author={Huang, Siming and Cheng, Tianhao and Liu, Jason Klein and Hao, Jiaran and Song, Liuyihan and Xu, Yang and Yang, J and Liu, JH and Zhang, Chenchen and Chai, Linzheng and others},
  journal={arXiv preprint arXiv:2411.04905},
  year={2024}
}

@misc{claude37,
  title = {Claude 3.7 Sonnet and Claude Code},
  author = {Anthropic},
  institution = {Anthropic},
  url = {https://www.anthropic.com/news/claude-3-7-sonnet},
  year={2025}
}

@misc{gpt45,
  title = {Introducing GPT-4.5},
  author = {OpenAI},
  institution = {OpenAI},
  url = {https://openai.com/index/introducing-gpt-4-5/},
  year={2025}
}

@article{starcoder,
  author       = {Raymond Li and
                  Loubna Ben Allal and
                  Yangtian Zi and
                  Niklas Muennighoff and
                  Denis Kocetkov and
                  Chenghao Mou and
                  Marc Marone and
                  Christopher Akiki and
                  Jia Li and
                  Jenny Chim and
                  Qian Liu and
                  Evgenii Zheltonozhskii and
                  Terry Yue Zhuo and
                  Thomas Wang and
                  Olivier Dehaene and
                  Mishig Davaadorj and
                  Joel Lamy{-}Poirier and
                  Jo{\~{a}}o Monteiro and
                  Oleh Shliazhko and
                  Nicolas Gontier and
                  Nicholas Meade and
                  Armel Zebaze and
                  Ming{-}Ho Yee and
                  Logesh Kumar Umapathi and
                  Jian Zhu and
                  Benjamin Lipkin and
                  Muhtasham Oblokulov and
                  Zhiruo Wang and
                  Rudra Murthy V and
                  Jason Stillerman and
                  Siva Sankalp Patel and
                  Dmitry Abulkhanov and
                  Marco Zocca and
                  Manan Dey and
                  Zhihan Zhang and
                  Nour Moustafa{-}Fahmy and
                  Urvashi Bhattacharyya and
                  Wenhao Yu and
                  Swayam Singh and
                  Sasha Luccioni and
                  Paulo Villegas and
                  Maxim Kunakov and
                  Fedor Zhdanov and
                  Manuel Romero and
                  Tony Lee and
                  Nadav Timor and
                  Jennifer Ding and
                  Claire Schlesinger and
                  Hailey Schoelkopf and
                  Jan Ebert and
                  Tri Dao and
                  Mayank Mishra and
                  Alex Gu and
                  Jennifer Robinson and
                  Carolyn Jane Anderson and
                  Brendan Dolan{-}Gavitt and
                  Danish Contractor and
                  Siva Reddy and
                  Daniel Fried and
                  Dzmitry Bahdanau and
                  Yacine Jernite and
                  Carlos Mu{\~{n}}oz Ferrandis and
                  Sean Hughes and
                  Thomas Wolf and
                  Arjun Guha and
                  Leandro von Werra and
                  Harm de Vries},
  title        = {StarCoder: may the source be with you!},
  journal      = {arXiv preprint arXiv:2305.06161},
  volume       = {abs/2305.06161},
  year         = {2023},
  url          = {https://arxiv.org/abs/2305.06161},
  doi          = {10.48550/arXiv.2305.06161},
  eprinttype    = {arXiv},
  eprint       = {2305.06161},
}

@article{mbpp,
  title={Program Synthesis with Large Language Models},
  author={Austin, Jacob and Odena, Augustus and Nye, Maxwell and Bosma, Maarten and Michalewski, Henryk and Dohan, David and Jiang, Ellen and Cai, Carrie and Terry, Michael and Le, Quoc and others},
  journal={arXiv preprint arXiv:2108.07732},
  year={2021},
  url={https://arxiv.org/abs/2108.07732}
}

@article{bloom,
  title={BLOOM: A 176B-Parameter Open-Access Multilingual Language Model},
  author={Scao, Teven Le and Fan, Angela and Akiki, Christopher and Pavlick, Ellie and Ili{\'c}, Suzana and Hesslow, Daniel and Castagn{\'e}, Roman and Luccioni, Alexandra Sasha and Yvon, Fran{\c{c}}ois and Gall{\'e}, Matthias and others},
  journal={arXiv preprint arXiv:2211.05100},
  year={2022},
  url={https://arxiv.org/abs/2211.05100},
}

@article{adamw,
  title={Decoupled Weight Decay Regularization},
  author={Loshchilov, Ilya and Hutter, Frank},
  journal={arXiv preprint arXiv:1711.05101},
  year={2017},
  url={https://arxiv.org/abs/1711.05101},
}

@article{gpt,
  title        = {Improving language understanding by generative pre-training},
  author       = {Radford, Alec and Narasimhan, Karthik and Salimans, Tim and Sutskever, Ilya and others},
  year={2018},
  publisher    = {OpenAI},
  journal      = {OpenAI blog},
  url          = {https://openai.com/research/language-unsupervised},
}

@article{AlphaCode,
  author       = {Yujia Li and
                  David H. Choi and
                  Junyoung Chung and
                  Nate Kushman and
                  Julian Schrittwieser and
                  R{\'{e}}mi Leblond and
                  Tom Eccles and
                  James Keeling and
                  Felix Gimeno and
                  Agustin Dal Lago and
                  Thomas Hubert and
                  Peter Choy and
                  Cyprien de Masson d'Autume and
                  Igor Babuschkin and
                  Xinyun Chen and
                  Po{-}Sen Huang and
                  Johannes Welbl and
                  Sven Gowal and
                  Alexey Cherepanov and
                  James Molloy and
                  Daniel J. Mankowitz and
                  Esme Sutherland Robson and
                  Pushmeet Kohli and
                  Nando de Freitas and
                  Koray Kavukcuoglu and
                  Oriol Vinyals},
  title        = {Competition-Level Code Generation with AlphaCode},
  journal      = {arXiv preprint arXiv:2203.07814},
  volume       = {abs/2203.07814},
  year         = {2022},
  url={https://arxiv.org/abs/2203.07814},
}

@article{codex,
  author       = {Mark Chen and
                  Jerry Tworek and
                  Heewoo Jun and
                  Qiming Yuan and
                  Henrique Pond{\'{e}} de Oliveira Pinto and
                  Jared Kaplan and
                  Harrison Edwards and
                  Yuri Burda and
                  Nicholas Joseph and
                  Greg Brockman and
                  Alex Ray and
                  Raul Puri and
                  Gretchen Krueger and
                  Michael Petrov and
                  Heidy Khlaaf and
                  Girish Sastry and
                  Pamela Mishkin and
                  Brooke Chan and
                  Scott Gray and
                  Nick Ryder and
                  Mikhail Pavlov and
                  Alethea Power and
                  Lukasz Kaiser and
                  Mohammad Bavarian and
                  Clemens Winter and
                  Philippe Tillet and
                  Felipe Petroski Such and
                  Dave Cummings and
                  Matthias Plappert and
                  Fotios Chantzis and
                  Elizabeth Barnes and
                  Ariel Herbert{-}Voss and
                  William Hebgen Guss and
                  Alex Nichol and
                  Alex Paino and
                  Nikolas Tezak and
                  Jie Tang and
                  Igor Babuschkin and
                  Suchir Balaji and
                  Shantanu Jain and
                  William Saunders and
                  Christopher Hesse and
                  Andrew N. Carr and
                  Jan Leike and
                  Joshua Achiam and
                  Vedant Misra and
                  Evan Morikawa and
                  Alec Radford and
                  Matthew Knight and
                  Miles Brundage and
                  Mira Murati and
                  Katie Mayer and
                  Peter Welinder and
                  Bob McGrew and
                  Dario Amodei and
                  Sam McCandlish and
                  Ilya Sutskever and
                  Wojciech Zaremba},
  title        = {Evaluating Large Language Models Trained on Code},
  journal      = {arXiv preprint arXiv:2107.03374},
  volume       = {abs/2107.03374},
  year         = {2021},
  url          = {https://arxiv.org/abs/2107.03374},
  eprinttype    = {arXiv},
  eprint       = {2107.03374},
}

@article{santacoder,
  title={{SantaCoder}: Don't reach for the stars!},
  author={Allal, Loubna Ben and Li, Raymond and Kocetkov, Denis and Mou, Chenghao and Akiki, Christopher and Ferrandis, Carlos Munoz and Muennighoff, Niklas and Mishra, Mayank and Gu, Alex and Dey, Manan and others},
  journal={arXiv preprint arXiv:2301.03988},
  year={2023},
  url={https://arxiv.org/abs/2301.03988}
}

@article{codet5,
  title={CodeT5: Identifier-aware Unified Pre-trained Encoder-Decoder Models for Code Understanding and Generation},
  author={Wang, Yue and Wang, Weishi and Joty, Shafiq and Hoi, Steven CH},
  journal={arXiv preprint arXiv:2109.00859},
  year={2021},
  url={https://arxiv.org/abs/2109.00859},
}

@article{livecodebench,
  title={LiveCodeBench: Holistic and Contamination Free Evaluation of Large Language Models for Code},
  author={Jain, Naman and Han, King and Gu, Alex and Li, Wen-Ding and Yan, Fanjia and Zhang, Tianjun and Wang, Sida and Solar-Lezama, Armando and Sen, Koushik and Stoica, Ion},
  journal={arXiv preprint arXiv:2403.07974},
  year={2024},
  url={https://arxiv.org/abs/2403.07974},
}

@inproceedings{bert,
  author       = {Jacob Devlin and
                  Ming{-}Wei Chang and
                  Kenton Lee and
                  Kristina Toutanova},
  title        = {{BERT:} Pre-training of Deep Bidirectional Transformers for Language
                  Understanding},
  booktitle    = {Proceedings of the 2019 Conference of the North American Chapter of
                  the Association for Computational Linguistics: Human Language Technologies,
                  {NAACL-HLT} 2019, Minneapolis, MN, USA, June 2-7, 2019, Volume 1 (Long
                  and Short Papers)},
  pages        = {4171--4186},
  publisher    = {Association for Computational Linguistics},
  year         = {2019},
  url          = {https://doi.org/10.18653/v1/n19-1423},
  doi          = {10.18653/V1/N19-1423},
}

@article{repocoder,
  author       = {Fengji Zhang and
                  Bei Chen and
                  Yue Zhang and
                  Jin Liu and
                  Daoguang Zan and
                  Yi Mao and
                  Jian{-}Guang Lou and
                  Weizhu Chen},
  title        = {{RepoCoder}: Repository-Level Code Completion Through Iterative Retrieval
                  and Generation},
  journal      = {arXiv preprint arXiv:2303.12570},
  volume       = {abs/2303.12570},
  year         = {2023},
  url          = {https://doi.org/10.48550/arXiv.2303.12570},
  doi          = {10.48550/arXiv.2303.12570},
  eprinttype    = {arXiv},
  eprint       = {2303.12570},
}

@inproceedings{ds1000,
  author       = {Yuhang Lai and
                  Chengxi Li and
                  Yiming Wang and
                  Tianyi Zhang and
                  Ruiqi Zhong and
                  Luke Zettlemoyer and
                  Wen{-}Tau Yih and
                  Daniel Fried and
                  Sida I. Wang and
                  Tao Yu},
  title        = {{DS-1000:} {A} Natural and Reliable Benchmark for Data Science Code
                  Generation},
  booktitle    = {International Conference on Machine Learning, {ICML} 2023, 23-29 July
                  2023, Honolulu, Hawaii, {USA}},
  series       = {Proceedings of Machine Learning Research},
  volume       = {202},
  pages        = {18319--18345},
  publisher    = {{PMLR}},
  year         = {2023},
  url          = {https://proceedings.mlr.press/v202/lai23b.html},
}

@article{codeelo,
  title={CodeElo: Benchmarking Competition-level Code Generation of LLMs with Human-comparable Elo Ratings},
  author={Quan, Shanghaoran and Yang, Jiaxi and Yu, Bowen and Zheng, Bo and Liu, Dayiheng and Yang, An and Ren, Xuancheng and Gao, Bofei and Miao, Yibo and Feng, Yunlong and others},
  journal={arXiv preprint arXiv:2501.01257},
  year={2025}
}

@article{starcoder2,
  title={StarCoder 2 and The Stack v2: The Next Generation},
  author={Lozhkov, Anton and Li, Raymond and Allal, Loubna Ben and Cassano, Federico and Lamy-Poirier, Joel and Tazi, Nouamane and Tang, Ao and Pykhtar, Dmytro and Liu, Jiawei and Wei, Yuxiang and others},
  journal={arXiv preprint arXiv:2402.19173},
  year={2024},
  url={https://arxiv.org/abs/2402.19173}
}

@article{multipl_e,
  title={MultiPL-E: A Scalable and Polyglot Approach to Benchmarking Neural Code Generation},
  author={Cassano, Federico and Gouwar, John and Nguyen, Daniel and Nguyen, Sydney and Phipps-Costin, Luna and Pinckney, Donald and Yee, Ming-Ho and Zi, Yangtian and Anderson, Carolyn Jane and Feldman, Molly Q and others},
  journal={IEEE Transactions on Software Engineering},
  year={2023},
  publisher={IEEE},
  url={https://ieeexplore.ieee.org/abstract/document/10103177},
}

@InProceedings{BabelCode,
  title = 	 {Measuring the Impact of Programming Language Distribution},
  author =       {Orlanski, Gabriel and Xiao, Kefan and Garcia, Xavier and Hui, Jeffrey and Howland, Joshua and Malmaud, Jonathan and Austin, Jacob and Singh, Rishabh and Catasta, Michele},
  booktitle = 	 {Proceedings of the 40th International Conference on Machine Learning},
  pages = 	 {26619--26645},
  year = 	 {2023},
  editor = 	 {Krause, Andreas and Brunskill, Emma and Cho, Kyunghyun and Engelhardt, Barbara and Sabato, Sivan and Scarlett, Jonathan},
  volume = 	 {202},
  series = 	 {Proceedings of Machine Learning Research},
  month = 	 {23--29 Jul},
  publisher =    {PMLR},
  url = 	 {https://proceedings.mlr.press/v202/orlanski23a.html},
}

@article{humaneval_xl,
  title={HumanEval-XL: A Multilingual Code Generation Benchmark for Cross-lingual Natural Language Generalization},
  author={Peng, Qiwei and Chai, Yekun and Li, Xuhong},
  journal={arXiv preprint arXiv:2402.16694},
  year={2024},
  url={https://arxiv.org/abs/2402.16694},
}

@article{humaneval_x,
  title={CodeGeeX: A Pre-Trained Model for Code Generation with Multilingual Evaluations on HumanEval-X},
  author={Zheng, Qinkai and Xia, Xiao and Zou, Xu and Dong, Yuxiao and Wang, Shan and Xue, Yufei and Wang, Zihan and Shen, Lei and Wang, Andi and Li, Yang and others},
  journal={arXiv preprint arXiv:2303.17568},
  year={2023},
  url={https://arxiv.org/abs/2303.17568},
}

@inproceedings{codereval,
  title={CoderEval: A Benchmark of Pragmatic Code Generation with Generative Pre-trained Models},
  author={Yu, Hao and Shen, Bo and Ran, Dezhi and Zhang, Jiaxin and Zhang, Qi and Ma, Yuchi and Liang, Guangtai and Li, Ying and Wang, Qianxiang and Xie, Tao},
  booktitle={Proceedings of the 46th IEEE/ACM International Conference on Software Engineering},
  pages={1--12},
  year={2024},
  url={https://dl.acm.org/doi/abs/10.1145/3597503.3623316},
}

@inproceedings{arcade_nl2code,
    title = "Natural Language to Code Generation in Interactive Data Science Notebooks",
    author = "Yin, Pengcheng  and
      Li, Wen-Ding  and
      Xiao, Kefan  and
      Rao, Abhishek  and
      Wen, Yeming  and
      Shi, Kensen  and
      Howland, Joshua  and
      Bailey, Paige  and
      Catasta, Michele  and
      Michalewski, Henryk  and
      Polozov, Oleksandr  and
      Sutton, Charles",
    editor = "Rogers, Anna  and
      Boyd-Graber, Jordan  and
      Okazaki, Naoaki",
    booktitle = "Proceedings of the 61st Annual Meeting of the Association for Computational Linguistics (Volume 1: Long Papers)",
    month = jul,
    year = "2023",
    address = "Toronto, Canada",
    publisher = "Association for Computational Linguistics",
    url = "https://aclanthology.org/2023.acl-long.9",
    doi = "10.18653/v1/2023.acl-long.9",
    pages = "126--173",
}

@article{xcodeeval,
  title={xCodeEval: A Large Scale Multilingual Multitask Benchmark for Code Understanding, Generation, Translation and Retrieval},
  author={Khan, Mohammad Abdullah Matin and Bari, M Saiful and Do, Xuan Long and Wang, Weishi and Parvez, Md Rizwan and Joty, Shafiq},
  journal={arXiv preprint arXiv:2303.03004},
  year={2023},
  url={https://arxiv.org/abs/2303.03004},
}

@article{debugbench,
  title={DebugBench: Evaluating Debugging Capability of Large Language Models},
  author={Tian, Runchu and Ye, Yining and Qin, Yujia and Cong, Xin and Lin, Yankai and Liu, Zhiyuan and Sun, Maosong},
  journal={arXiv preprint arXiv:2401.04621},
  year={2024},
  url={https://arxiv.org/abs/2401.04621},
}

@inproceedings{quixbugs,
  title={QuixBugs: a multi-lingual program repair benchmark set based on the quixey challenge},
  author={Lin, Derrick and Koppel, James and Chen, Angela and Solar-Lezama, Armando},
  booktitle={Proceedings Companion of the 2017 ACM SIGPLAN international conference on systems, programming, languages, and applications: software for humanity},
  pages={55--56},
  year={2017},
  url={https://dl.acm.org/doi/abs/10.1145/3135932.3135941},
}

@article{runbugrun,
  title={RunBugRun -- An Executable Dataset for Automated Program Repair},
  author={Prenner, Julian Aron and Robbes, Romain},
  journal={arXiv preprint arXiv:2304.01102},
  year={2023},
  url={https://arxiv.org/abs/2304.01102},
}

@article{EvalGPTFix,
  title={A Critical Review of Large Language Model on Software Engineering: An Example from ChatGPT and Automated Program Repair},
  author={Zhang, Quanjun and Zhang, Tongke and Zhai, Juan and Fang, Chunrong and Yu, Bowen and Sun, Weisong and Chen, Zhenyu},
  journal={arXiv preprint arXiv:2310.08879},
  year={2023},
  url={https://arxiv.org/abs/2310.08879},
}

@misc{llama3,
  author       = "Meta AI",
  title        = "Introducing Meta Llama 3: The most capable openly available LLM to date",
  howpublished = "\url{https://ai.meta.com/blog/meta-llama-3/}",
  month        = "apr",
  year         = "2024",
}

@article{codegemma,
  title={Gemma: Open Models Based on Gemini Research and Technology},
  author={Gemma Team, Google},
  journal={arXiv preprint arXiv:2403.08295},
  year={2024},
  url={https://arxiv.org/abs/2403.08295},
}

@article{swe_bench,
  title={SWE-bench: Can Language Models Resolve Real-World GitHub Issues?},
  author={Jimenez, Carlos E and Yang, John and Wettig, Alexander and Yao, Shunyu and Pei, Kexin and Press, Ofir and Narasimhan, Karthik},
  journal={arXiv preprint arXiv:2310.06770},
  year={2023},
  url={https://arxiv.org/abs/2310.06770},
}

@article{naturalcodebench,
  title={NaturalCodeBench: Examining Coding Performance Mismatch on HumanEval and Natural User Prompts},
  author={Zhang, Shudan and Zhao, Hanlin and Liu, Xiao and Zheng, Qinkai and Qi, Zehan and Gu, Xiaotao and Zhang, Xiaohan and Dong, Yuxiao and Tang, Jie},
  journal={arXiv preprint arXiv:2405.04520},
  year={2024}
}

@article{mceval,
  title={McEval: Massively Multilingual Code Evaluation},
  author={Chai, Linzheng and Liu, Shukai and Yang, Jian and Yin, Yuwei and Jin, Ke and Liu, Jiaheng and Sun, Tao and Zhang, Ge and Ren, Changyu and Guo, Hongcheng and others},
  journal={arXiv preprint arXiv:2406.07436},
  year={2024}
}

@article{bigcodebench,
  title={Bigcodebench: Benchmarking code generation with diverse function calls and complex instructions},
  author={Zhuo, Terry Yue and Vu, Minh Chien and Chim, Jenny and Hu, Han and Yu, Wenhao and Widyasari, Ratnadira and Yusuf, Imam Nur Bani and Zhan, Haolan and He, Junda and Paul, Indraneil and others},
  journal={arXiv preprint arXiv:2406.15877},
  year={2024}
}

@inproceedings{unicoder,
    title = "{U}ni{C}oder: Scaling Code Large Language Model via Universal Code",
    author = "Sun, Tao  and
      Chai, Linzheng  and
      Yang, Jian  and
      Yin, Yuwei  and
      Guo, Hongcheng  and
      Liu, Jiaheng  and
      Wang, Bing  and
      Yang, Liqun  and
      Li, Zhoujun",
    editor = "Ku, Lun-Wei  and
      Martins, Andre  and
      Srikumar, Vivek",
    booktitle = "Proceedings of the 62nd Annual Meeting of the Association for Computational Linguistics (Volume 1: Long Papers)",
    month = aug,
    year = "2024",
    address = "Bangkok, Thailand",
    publisher = "Association for Computational Linguistics",
    url = "https://aclanthology.org/2024.acl-long.100",
    pages = "1812--1824",
}

@article{mammoth2,
  title={Mammoth2: Scaling instructions from the web},
  author={Yue, Xiang and Zheng, Tuney and Zhang, Ge and Chen, Wenhu},
  journal={arXiv preprint arXiv:2405.03548},
  year={2024}
}

@article{kun,
  title={Kun: Answer Polishment for Chinese Self-Alignment with Instruction Back-Translation},
  author={Zheng, Tianyu and Guo, Shuyue and Qu, Xingwei and Guo, Jiawei and Zhang, Weixu and Du, Xinrun and Lin, Chenghua and Huang, Wenhao and Chen, Wenhu and Fu, Jie and others},
  journal={arXiv preprint arXiv:2401.06477},
  year={2024}
}

@article{qwen_coder,
  title={Qwen2. 5-Coder Technical Report},
  author={Hui, Binyuan and Yang, Jian and Cui, Zeyu and Yang, Jiaxi and Liu, Dayiheng and Zhang, Lei and Liu, Tianyu and Zhang, Jiajun and Yu, Bowen and Dang, Kai and others},
  journal={arXiv preprint arXiv:2409.12186},
  year={2024}
}

@article{debugeval,
  title={Enhancing the Code Debugging Ability of LLMs via Communicative Agent Based Data Refinement},
  author={Yang, Weiqing and Wang, Hanbin and Liu, Zhenghao and Li, Xinze and Yan, Yukun and Wang, Shuo and Gu, Yu and Yu, Minghe and Liu, Zhiyuan and Yu, Ge},
  journal={arXiv preprint arXiv:2408.05006},
  year={2024}
}

@article{coder,
  title={CodeR: Issue Resolving with Multi-Agent and Task Graphs},
  author={Chen, Dong and Lin, Shaoxin and Zeng, Muhan and Zan, Daoguang and Wang, Jian-Gang and Cheshkov, Anton and Sun, Jun and Yu, Hao and Dong, Guoliang and Aliev, Artem and others},
  journal={arXiv preprint arXiv:2406.01304},
  year={2024}
}

@article{magis,
  title={MAGIS: LLM-Based Multi-Agent Framework for GitHub Issue Resolution},
  author={Tao, Wei and Zhou, Yucheng and Zhang, Wenqiang and Cheng, Yu},
  journal={arXiv preprint arXiv:2403.17927},
  year={2024}

}

@article{zhong2024advancing,
  title={Advancing Bug Detection in Fastjson2 with Large Language Models Driven Unit Test Generation},
  author={Zhong, Zhiyuan and Wang, Sinan and Wang, Hailong and Wen, Shaojin and Guan, Hao and Tao, Yida and Liu, Yepang},
  journal={arXiv preprint arXiv:2410.09414},
  year={2024}
}

@article{pradel2018deepbugs,
  title={Deepbugs: A learning approach to name-based bug detection},
  author={Pradel, Michael and Sen, Koushik},
  journal={Proceedings of the ACM on Programming Languages},
  volume={2},
  number={OOPSLA},
  pages={1--25},
  year={2018},
  publisher={ACM New York, NY, USA}
}

@article{allamanis2021self,
  title={Self-supervised bug detection and repair},
  author={Allamanis, Miltiadis and Jackson-Flux, Henry and Brockschmidt, Marc},
  journal={Advances in Neural Information Processing Systems},
  volume={34},
  pages={27865--27876},
  year={2021}
}

@article{yuan2023evaluating,
  title={Evaluating instruction-tuned large language models on code comprehension and generation},
  author={Yuan, Zhiqiang and Liu, Junwei and Zi, Qiancheng and Liu, Mingwei and Peng, Xin and Lou, Yiling},
  journal={arXiv preprint arXiv:2308.01240},
  year={2023}
}

@inproceedings{zhang2024prompt,
  title={Prompt-enhanced software vulnerability detection using chatgpt},
  author={Zhang, Chenyuan and Liu, Hao and Zeng, Jiutian and Yang, Kejing and Li, Yuhong and Li, Hui},
  booktitle={Proceedings of the 2024 IEEE/ACM 46th International Conference on Software Engineering: Companion Proceedings},
  pages={276--277},
  year={2024}
}

@article{yang2024fuzzcoder,
  title={FuzzCoder: Byte-level Fuzzing Test via Large Language Model},
  author={Yang, Liqun and Yang, Jian and Wei, Chaoren and Niu, Guanglin and Zhang, Ge and Wang, Yunli and ChaI, Linzheng and Xia, Wanxu and Guo, Hongcheng and Zhang, Shun and others},
  journal={arXiv preprint arXiv:2409.01944},
  year={2024}
}

@inproceedings{xia2024fuzz4all,
  title={Fuzz4all: Universal fuzzing with large language models},
  author={Xia, Chunqiu Steven and Paltenghi, Matteo and Le Tian, Jia and Pradel, Michael and Zhang, Lingming},
  booktitle={Proceedings of the IEEE/ACM 46th International Conference on Software Engineering},
  pages={1--13},
  year={2024}
}

@inproceedings{deng2023large,
  title={Large language models are zero-shot fuzzers: Fuzzing deep-learning libraries via large language models},
  author={Deng, Yinlin and Xia, Chunqiu Steven and Peng, Haoran and Yang, Chenyuan and Zhang, Lingming},
  booktitle={Proceedings of the 32nd ACM SIGSOFT international symposium on software testing and analysis},
  pages={423--435},
  year={2023}
}

@article{wen2024fixing,
  title={Fixing code generation errors for large language models},
  author={Wen, Hao and Zhu, Yueheng and Liu, Chao and Ren, Xiaoxue and Du, Weiwei and Yan, Meng},
  journal={arXiv preprint arXiv:2409.00676},
  year={2024}
}

@article{gu2024counterfeit,
  title={The counterfeit conundrum: Can code language models grasp the nuances of their incorrect generations?},
  author={Gu, Alex and Li, Wen-Ding and Jain, Naman and Olausson, Theo X and Lee, Celine and Sen, Koushik and Solar-Lezama, Armando},
  journal={arXiv preprint arXiv:2402.19475},
  year={2024}
}

@article{tambon2024bugs,
  title={Bugs in large language models generated code: An empirical study},
  author={Tambon, Florian and Dakhel, Arghavan Moradi and Nikanjam, Amin and Khomh, Foutse and Desmarais, Michel C and Antoniol, Giuliano},
  journal={CoRR},
  year={2024}
}

@article{wang2024large,
  title={Where Do Large Language Models Fail When Generating Code?},
  author={Wang, Zhijie and Zhou, Zijie and Song, Da and Huang, Yuheng and Chen, Shengmai and Ma, Lei and Zhang, Tianyi},
  journal={arXiv preprint arXiv:2406.08731},
  year={2024}
}

@inproceedings{sobania2023analysis,
  title={An analysis of the automatic bug fixing performance of chatgpt},
  author={Sobania, Dominik and Briesch, Martin and Hanna, Carol and Petke, Justyna},
  booktitle={2023 IEEE/ACM International Workshop on Automated Program Repair (APR)},
  pages={23--30},
  year={2023},
  organization={IEEE}
}

@article{xia2023conversational,
  title={Conversational automated program repair},
  author={Xia, Chunqiu Steven and Zhang, Lingming},
  journal={arXiv preprint arXiv:2301.13246},
  year={2023}
}

@inproceedings{prenner2022can,
  title={Can OpenAI's codex fix bugs? an evaluation on QuixBugs},
  author={Prenner, Julian Aron and Babii, Hlib and Robbes, Romain},
  booktitle={Proceedings of the Third International Workshop on Automated Program Repair},
  pages={69--75},
  year={2022}
}

@article{ifeval,
  title={Instruction-following evaluation for large language models},
  author={Zhou, Jeffrey and Lu, Tianjian and Mishra, Swaroop and Brahma, Siddhartha and Basu, Sujoy and Luan, Yi and Zhou, Denny and Hou, Le},
  journal={arXiv preprint arXiv:2311.07911},
  year={2023}
}

@article{codearena,
  title={Evaluating and aligning codellms on human preference},
  author={Yang, Jian and Yang, Jiaxi and Jin, Ke and Miao, Yibo and Zhang, Lei and Yang, Liqun and Cui, Zeyu and Zhang, Yichang and Hui, Binyuan and Lin, Junyang},
  journal={arXiv preprint arXiv:2412.05210},
  year={2024}
}

@article{llm_as_a_judge,
  title={Judging llm-as-a-judge with mt-bench and chatbot arena},
  author={Zheng, Lianmin and Chiang, Wei-Lin and Sheng, Ying and Zhuang, Siyuan and Wu, Zhanghao and Zhuang, Yonghao and Lin, Zi and Li, Zhuohan and Li, Dacheng and Xing, Eric and others},
  journal={Advances in Neural Information Processing Systems},
  volume={36},
  pages={46595--46623},
  year={2023}
}

@article{mdeval,
  title={Mdeval: Massively multilingual code debugging},
  author={Liu, Shukai and Chai, Linzheng and Yang, Jian and Shi, Jiajun and Zhu, He and Wang, Liran and Jin, Ke and Zhang, Wei and Zhu, Hualei and Guo, Shuyue and others},
  journal={arXiv preprint arXiv:2411.02310},
  year={2024}
}

@article{fullstackbench,
  title={Fullstack bench: Evaluating llms as full stack coder},
  author={Liu, Siyao and Zhu, He and Liu, Jerry and Xin, Shulin and Li, Aoyan and Long, Rui and Chen, Li and Yang, Jack and Xia, Jinxiang and Peng, ZY and others},
  journal={arXiv preprint arXiv:2412.00535},
  year={2024}
}

@article{o1_mini,
  title={Openai o1 system card},
  author={Jaech, Aaron and Kalai, Adam and Lerer, Adam and Richardson, Adam and El-Kishky, Ahmed and Low, Aiden and Helyar, Alec and Madry, Aleksander and Beutel, Alex and Carney, Alex and others},
  journal={arXiv preprint arXiv:2412.16720},
  year={2024}
}

@article{deepseekv3,
  title={Deepseek-v3 technical report},
  author={Liu, Aixin and Feng, Bei and Xue, Bing and Wang, Bingxuan and Wu, Bochao and Lu, Chengda and Zhao, Chenggang and Deng, Chengqi and Zhang, Chenyu and Ruan, Chong and others},
  journal={arXiv preprint arXiv:2412.19437},
  year={2024}
}

@article{deepseek_r1,
  title={Deepseek-r1: Incentivizing reasoning capability in llms via reinforcement learning},
  author={Guo, Daya and Yang, Dejian and Zhang, Haowei and Song, Junxiao and Zhang, Ruoyu and Xu, Runxin and Zhu, Qihao and Ma, Shirong and Wang, Peiyi and Bi, Xiao and others},
  journal={arXiv preprint arXiv:2501.12948},
  year={2025}
}
